\pdfoutput=1
\documentclass[11pt]{article}

\usepackage[]{ACL2023}

\usepackage{times}
\usepackage{latexsym}

\usepackage{amsmath}

\usepackage[T1]{fontenc}
\usepackage[utf8]{inputenc}

\usepackage{microtype}

\usepackage{inconsolata}

\usepackage{hyperref}

\usepackage{multirow}
\usepackage{graphicx}
\usepackage{colortbl} %彩色表格需要加载的宏包
\usepackage{xcolor}
\definecolor{color1}{RGB}{68,115,169}
\definecolor{color2}{RGB}{188,215,237}
\definecolor{color3}{RGB}{254,190,0}
\definecolor{color4}{RGB}{254,0,0}
\usepackage{float}

\usepackage{longtable}
\usepackage{enumitem}

\usepackage{booktabs}

\usepackage[normalem]{ulem}
\useunder{\uline}{\ul}{}

\newcommand{\hide}[1]{} %hide
\newcommand{\vpara}[1]{\vspace{0.05in}\noindent \textbf{#1 }}

\newcommand{\model}{CoT-KA}
\newcommand{\smodel}{CoT-KA }

\title{Chain of Thought Prompting Elicits Knowledge Augmentation}

\author{
    Dingjun Wu$^1$, Jing Zhang$^{2}$ \thanks{\ \  Corresponding author: Jing Zhang.}, Xinmei Huang$^2$ \\ 
    $^1$Tsinghua Shenzhen International Graduate School, Tsinghua University \\ 
    $^2$School of Information, Renmin University of China \\ 
    \texttt{wudj20@mails.tsinghua.edu.cn} \\
    \texttt{\{zhang-jing,huangxinmei\}@ruc.edu.cn} \\
}

\begin{document}
\maketitle
\begin{abstract}
% The knowledge-augmented deep learning paradigm refers to the deep learning paradigm in which domain knowledge is identified and integrated into deep models. Conventional methods create a task-specific method to gather external knowledge from various sources. Large language models, on the other hand, are well pre-trained and can be used as a general source of external knowledge. 

% In this paper, we propose \model, a Chain-of-Thought-based method to augment knowledge for deep learning, which avoids the extra knowledge retrieval or knowledge reasoning model as in the conventional augmentation methods. \smodel outperforms both the pure CoT-based methods and the conventional knowledge augmentation techniques on the majority of the eleven publicly available benchmarks for various reasoning tasks. \footnote{Our code and data are available at \url{https://github.com/Alphonse-7/CoT-KA}}

The knowledge-augmented deep learning paradigm refers to a paradigm in which domain knowledge is identified and integrated into deep models. Conventional methods typically employ task-specific approaches to gather external knowledge from various sources. In contrast, large language models are extensively pre-trained and can serve as a comprehensive source of external knowledge. 
In this paper, we propose \model, a Chain-of-Thought-based method that augments knowledge for deep learning. CoT-KA avoids the need for additional knowledge retrieval or knowledge reasoning models, as required in conventional augmentation methods. Our results demonstrate that CoT-KA outperforms both pure CoT-based methods and the non-augmented method across the majority of eleven publicly available benchmarks for various reasoning tasks \footnote{Our code and data are available at \url{https://github.com/RUCKBReasoning/CoT-KA}}.

% The knowledge-augmented deep learning paradigm refers to a paradigm in which domain knowledge is identified and integrated into deep models. Conventional methods typically employ task-specific approaches to gather external knowledge from various sources. In contrast, large language models are extensively pre-trained and can serve as a comprehensive source of external knowledge. In this paper, we propose CoT-KA, a Chain-of-Thought-based method that augments knowledge for deep learning. CoT-KA avoids the need for additional knowledge retrieval or knowledge reasoning models, as required in conventional augmentation methods. Our results demonstrate that CoT-KA outperforms both pure CoT-based methods and the non-augmented method across the majority of eleven publicly available benchmarks for various reasoning tasks. 
\end{abstract}

\section{Introduction}
The Knowledge-Augmented deep learning (KADL) \cite{knowledge-augmanted-survey} paradigm refers to the deep learning paradigm in which domain knowledge is identified and integrated into the deep model. Adding domain knowledge makes it possible to develop deep learning that is data-efficient, generalizable, and interpretable ~\cite{knowledge-augmanted-survey}. 
% For example, retrieving external knowledge from Wikipedia is typically required for open domain question answering~\cite{FiD}. 
For example, retrieving external knowledge from an external knowledge pool like Wikipedia is typically required for open domain question answering and dialog generation ~\cite{FiD,zhang2023glm}. 
Logical equivalence laws such as contraposition and transitive laws help extend the implicit logical information \cite{reclor-dataset, LReasoner}.

\begin{figure}[ht]
    \centering
    \includegraphics[width=1.0\linewidth]{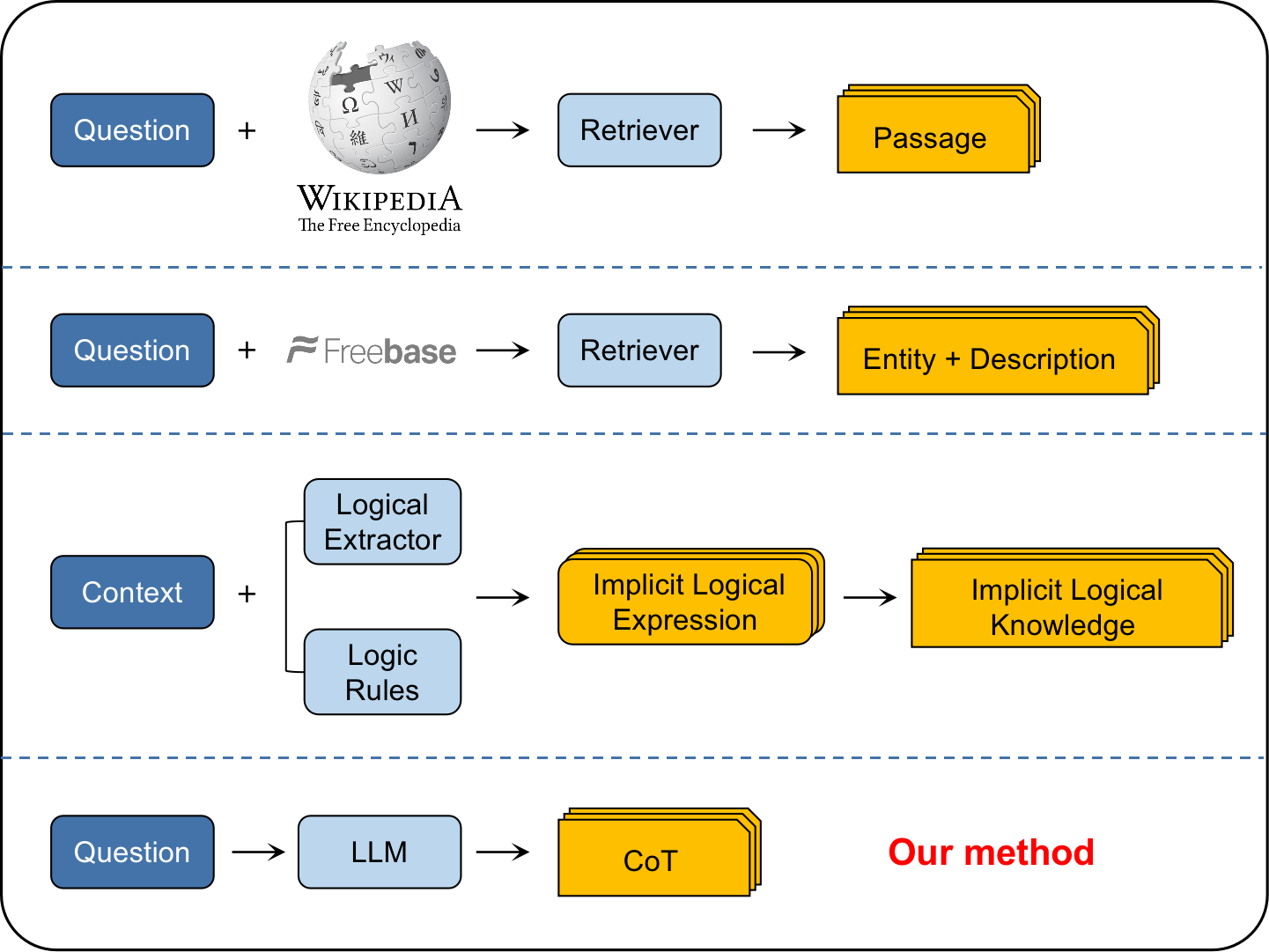}
    \caption{A various sources of external knowledge. We use LLM as our source of knowledge.}
    \label{fig1:knowledge-source}
\end{figure}

% External knowledge comes from various sources. For example, commonsense knowledge can be extracted from commonsense knowledge bases such as ConceptNet \cite{conceptnet} and ATOMIC \cite{atomic-knowledge-graph}, domain knowledge can be retrieved from knowledge bases such as Wikipedia and Freebase \cite{Freebase}, logic knowledge can be human-defined propositional or first-order logic, which is then created as rules for reasoning. In summary, existing knowledge augmentation methods require either creating a retriever to gather useful knowledge or developing a reasoner to take advantage of the logic rules in the external knowledge source.

External knowledge is derived from various sources. For instance, commonsense knowledge can be extracted from commonsense knowledge bases like ConceptNet \cite{conceptnet} and ATOMIC \cite{atomic-knowledge-graph}. Domain-specific knowledge can be retrieved from knowledge bases such as Wikipedia and Freebase \cite{Freebase}. Logic knowledge, on the other hand, can be in the form of human-defined propositional or first-order logic, which is then utilized as rules for reasoning. In summary, existing knowledge augmentation methods typically involve either creating a retriever to gather relevant knowledge or developing a reasoner to leverage the logical rules within the external knowledge sources ~\cite{openqa-wikipedia,FiD,LReasoner,zhang2023glm}.

Recently, large language models (LLMs) \cite{zhao2023survey} have shown their potential as both the source and the retriever or reasoner of external knowledge. LLMs are pre-trained on a huge scale of datasets. Thus, they have already embedded a large amount of knowledge into their parameters, which can be considered a source of external knowledge. The reasoning ability of LLMs allows them to provide knowledge from their parameters without needing an extra retriever or a reasoner. The latest chain-of-thought (CoT) prompting technique \cite{fewshot-cot}, which elicits LLMs to generate a series of sentences that mimic the reasoning process for arriving at the answers, improves the reasoning ability of LLMs. It has proved to be remarkably effective in a variety of complex reasoning tasks such as math word problems and commonsense question answering \cite{fewshot-cot}. CoT prompting shows potential as a general technique to retrieve knowledge from LLMs. 

% In this paper, we propose \model---a \underline{CoT}-based method to retrieve knowledge from LLMs for \underline{K}nowledge-\underline{A}ugmented deep learning. An LLM is used by \smodel as a knowledge source, and through CoT prompting, the LLM is guided to provide knowledge that can be used as evidence to support the downstream reasoning from the input to the answer. Unlike conventional KADL, \smodel avoids the extra knowledge retrieval or knowledge reasoning model. Specifically, we first extract CoTs as knowledge from an LLM by the few-shot  \cite{fewshot-cot} or zero-shot oriented CoT prompting \cite{zeroshot-cot}. The former requires writing a few demonstrations to tell LLM the intention of reasoning whereas the latter inspires the LLM by using a template like ``let's think step by step''. The extracted CoTs are then appended to the original input as the augmented text, which is denoted by a special token. Finally, we fine-tune the task-relevant small pre-trained language model on the train data with the inputs augmented by the CoTs from the LLM.

In this paper, we propose \smodel -- a \underline{CoT}-based method to retrieve knowledge from LLMs for \underline{K}nowledge-\underline{A}ugmented deep learning. \smodel utilizes an LLM as a knowledge source, leveraging CoT prompting to guide the LLM in providing knowledge that can serve as evidence to support downstream reasoning from the input to the answer. Unlike conventional KADL approaches, \smodel eliminates the need for additional knowledge retrieval or a separate knowledge reasoning model. 
Specifically, we begin by extracting CoTs as knowledge from the LLM using either few-shot \cite{fewshot-cot} or zero-shot \cite{zeroshot-cot} CoT prompting. The former involves providing a few demonstrations to guide the LLM's reasoning, while the latter employs a template such as ``let's think step by step'' to inspire the LLM. The extracted CoTs are then appended to the original inputs, marked by a special token, to create augmented text. 
% Finally, we fine-tune a task-relevant small pre-trained language model (PLM) on the training data using the inputs augmented with CoTs from the LLM. 
% Finally, we fine-tune a task-relevant small pre-trained language model (PLM) on the augmented dataset.
Finally, we fine-tune a small task-relevant pre-trained language model (PLM) on the dataset augmented with CoTs.
% Through this process, CoT-KA enables the integration of knowledge from the LLM into the deep learning model without the need for additional retrieval or reasoning components. This approach holds promise for enhancing Knowledge-Augmented deep learning tasks.

% pre-version
% Since the reasoning path and answer in CoT may be wrong, it is more appropriate to use CoT as an auxiliary knowledge for model learning than to parse from CoT directly to get the answers to the input questions.

% We generate CoTs using the public GPT-3 \cite{gpt3} (175B parameters) API\footnote{Public API available at \url{https://openai.com/api/}}, and take ALBERT ~\cite{albert} and DeBERTa ~\cite{debertav3} as the task-relevant model for NLU tasks. For NLG tasks, we adopt T5 \cite{T5} as the task-relevant model.
We generate CoTs using the public GPT-3 \cite{gpt3} (175B parameters) API\footnote{Public API available at \url{https://openai.com/api/}}. For NLU (Natural Language Understanding) tasks, we employ ALBERT \cite{albert} and DeBERTa \cite{debertav3} as the task-relevant models. T5 \cite{T5} is utilized as the task-relevant model for NLG (Natural Language Generation) tasks.
We evaluate models' performance using eleven benchmarks, including (i) commonsense reasoning (CSQA ~\cite{CSQA}, StrategyQA ~\cite{strategyQA}, Date Understanding, Sports Understanding ~\cite{big-bench}); (ii) arithmetic reasoning (AQUA-RAT \cite{aqua}, GSM8K \cite{gsm8k}, SVAMP \cite{svamp}, MultiArith \cite{multiarith}, SingleEq \cite{singleeq}, AddSub \cite{addsub}); (iii) symbolic reasoning (Last Letter Concatenation \cite{fewshot-cot}), where all commonsense reasoning benchmarks and AQUA-RAT are formulated as NLU tasks, and the other arithmetic reasoning benchmarks and Last Letter Concatenation are formulated as NLG tasks in this paper. Particularly, we convert all of the multi-choice question answering tasks into NLU tasks.
Extensive experimental results show that in the majority of tasks, \smodel outperforms the original fine-tuning results without the use of CoTs as augmented knowledge. \smodel also surpasses Few-Shot-CoT and Zero-Shot-CoT on LLMs, which directly parse answers from the generated CoTs.

\section{Related Work}

\paragraph{Knowledge Augmented Technology.}
% The integration of external knowledge into deep learning models via knowledge augmentation approaches have received a lot of attention in many NLP tasks such as question answering~\cite{openqa-wikipedia,FiD}, text classification~\cite{knowledge-augmanted-survey}, and logical reasoning~\cite{LReasoner}.
% For example, when answering open-domain questions, evidences for reasoning the answers are not provided as input \cite{FiD}. Thus, \citet{openqa-wikipedia} employ bigram hashing and TF-IDF matching to retrieve relevant documents from external knowledge sources, whereas Fusion-in-Decoder \cite{FiD} uses BM25 \cite{BM25} and DPR \cite{DPR} for the retrieval of evidences. As the augmentation to the questions, these evidences help to reason the answers.
% Another example is logic reasoning, a challenging task that requires a logic-level analysis of a given text to arrive at the correct answer. To perform such logic-level analysis, the human-defined logic rules need to be injected. \citet{LReasoner} propose LReasoner, a Logic-driven context extension framework, which extends implicit logical information by performing logical reasoning using human-defined logic rules over the explicitly embedded logical information in the original input. The implicit logical information is verbalized and concatenated with original input for the subsequent answer reasoning.
The integration of external knowledge into deep learning models through knowledge augmentation approaches has gained significant attention in various NLP tasks, including question answering ~\cite{openqa-wikipedia,FiD}, dialogue generation ~\cite{zhang2023glm}, and logical reasoning ~\cite{LReasoner}. 
For instance, in the context of answering open-domain questions where supporting evidence is not explicitly provided \cite{FiD}, \citet{openqa-wikipedia} utilized techniques such as bigram hashing and TF-IDF matching to retrieve relevant documents from external knowledge sources. 
Similarly, Fusion-in-Decoder \cite{FiD} employed methods like BM25 \cite{BM25} and DPR \cite{DPR} for evidence retrieval. By augmenting the questions with these retrieved pieces of evidence, the models can better reason and provide answers.
Logic reasoning is another challenging task that requires a deep understanding of the logical structure within a given text to arrive at the correct answer. To facilitate such logic-level analysis, human-defined logic rules are introduced. \citet{LReasoner} proposed LReasoner, a logic-driven context extension framework that extends implicit logical information by performing logical reasoning using these predefined rules. The framework enhances the original input by verbalizing and concatenating the implicit logical information, enabling subsequent answer reasoning.

% Fusion-in-Decoder and LReasoner inspire our work to extend the external knowledge into the original input. However, the knowledge in these knowledge augmentation methods comes from external knowledge bases or pre-defined logical rules, and the retriever for extracting the knowledge or the reasoner for employing the rules is necessary in this process. In contrast, we employ LLMs which do not need an extra retriever or reasoner to obtain the knowledge for augmentation. 
Fusion-in-Decoder and LReasoner inspire our work to extend the external knowledge into the original input. However, the knowledge in these knowledge augmentation methods is sourced from external knowledge bases or pre-defined logical rules, requiring a retriever for knowledge extraction or a reasoner for rule application in the process. In contrast, we utilize LLMs that eliminate the need for an additional retriever or reasoner to acquire knowledge for augmentation.
% our work is a neural-based method rather than a rule-based one, which extends the original text by a large language model without the need to design complex extension rules manually.

\paragraph{Chain of Thought Prompting on LLMs.}

A CoT is a series of intermediate natural language reasoning steps that lead to the final output, inspired by how humans use a deliberate thinking process to perform complicated tasks.
% Experiments on various LLMs such as GPT-3~\cite{gpt3} and PaLM~\cite{palm} show that CoT prompting improves performance on a variety of arithmetic, commonsense, and symbolic reasoning tasks \cite{fewshot-cot}. 
Experimental results using various LLMs, such as GPT-3 ~\cite{gpt3} and PaLM ~\cite{palm}, demonstrate that CoT prompting enhances performance across a range of arithmetic, commonsense, and symbolic reasoning tasks \cite{fewshot-cot}.

% \citet{fewshot-cot} first propose Few-Shot-CoT, in which manual design of a few demonstrations is required to carry out the generation of reasoning paths. In contrast to Few-Shot-CoT prompting that requires manually crafted demonstrations for various tasks, \citet{zeroshot-cot} propose Zero-Shot-CoT, which issues a single zero-shot prompt that elicits CoTs from LLMs. By simply adding ``Let's think step by step'' before each answer, Zero-Shot-CoT demonstrates that LLMs are capable zero-shot reasoners without the need for any manually constructed few-shot examples. \citet{self-consistency} propose a new decoding strategy, self-consistency, to sample multiple LLM outputs and then take a majority vote over them in order to make the model take into account many CoTs when generating answers. However, to achieve optimal performance, it must generate a large number of reasoning paths (for instance, 40 paths), which raises the cost of computing.
\citet{fewshot-cot} initially propose Few-Shot-CoT, which requires the manual design of a few demonstrations to facilitate the generation of reasoning paths. 
% In contrast to Few-Shot-CoT, which necessitates manually crafted demonstrations for various tasks, \citet{zeroshot-cot} propose Zero-Shot-CoT. This approach employs a single zero-shot prompt that elicits CoTs from LLMs. 
In contrast, \citet{zeroshot-cot} propose Zero-Shot-CoT, which employs a single zero-shot prompt that elicits CoTs from LLMs. 
By simply adding ``Let's think step by step'' before each answer, Zero-Shot-CoT demonstrates that LLMs are capable zero-shot reasoners without the need for any manually constructed few-shot examples. Furthermore, \citet{self-consistency} introduce a new decoding strategy called self-consistency, which involves sampling multiple LLM outputs and aggregating them through majority voting. This strategy encourages the model to consider multiple CoTs when generating answers. However, to achieve optimal performance, a large number of reasoning paths (e.g., 40 paths) must be generated, leading to increased computational costs.

% Studies have shown that LLM can perform CoT reasoning with  (Few-Shot-CoT), or a single zero-shot prompt such as \emph{Let's think step by step} (Zero-Shot-CoT). 
% Wei et al first explored Few-Shot-CoT prompting for eliciting multi-step reasoning behavior in large language models, Few-Shot-CoT prompting achieves stronger performance than standard prompting by eliciting the CoT reasoning ability with effective manual demonstrations. 
% Inspired by Zero-Shot-CoT and Few-Shot-CoT, \citet{auto-cot} proposed Automatic Chain-of-Thought Prompting (Auto-CoT) to automatically construct demonstrations, it samples questions with diversity and generates reasoning chains to construct demonstrations. However, these methods consider a single CoT to get the final answer, which leads to a certain randomness in model reasoning. 

% These CoT prompting methods all parse the answer directly from the CoTs. In contrast, in our method, these generated CoTs only serve as additional knowledge to enhance fine-tuning the task-relevant models. Additionally, our method performs well even when only a limited number of CoTs are provided, in contrast to self-consistency, which depends on generating a large number of CoTs.
All of these CoT prompting methods directly extract the answer from the CoTs. In contrast, our method utilizes the generated CoTs as supplementary knowledge to improve the fine-tuning of task-relevant models. Moreover, our method demonstrates good performance even when a limited number of CoTs are provided, unlike self-consistency, which relies on generating a large number of CoTs.

\section{Pilot Study}
% In this section, we simply append one CoT to the original input to investigate the validity of CoT-augmented fine-tuning on two commonsense reasoning datasets, CSQA and StrategyQA.
In this section, we explore the effectiveness of CoT-augmented fine-tuning by simply appending one CoT to the original input. We assess the validity of this approach on two commonsense reasoning datasets, CSQA and StrategyQA.

% enhance the fine-tuning of reasoning task.
% text extension can enhance reasoning task based on fine-tuning. As reported in LReasoner, using a logic-driven context extension framework to extend the logical expressions and verbalize these logical expressions to original input text, enhancing the performance of fine-tuning results on ReClor(a complex reasoning task requiring logical reasoning). It shows the potential of text extension for reasoning tasks. CoT prompting contains potential reasoning steps to solve the reasoning task, which is similar to Logic Extension in LReasoner, both of which contain potential logic information in the input text.

\vpara{CoT-augmented Fine-tuning.}
% We added a CoT to the original input text to obtain the extended input for fine-tuning ALBERT.  We employ ALBERT-large-v2.
% Specifically, we produce CoT using both few-shot and zero-shot CoT methods, also known as Few-Shot-CoT and Zero-Shot-CoT.
% Few-Shot-CoT employs the same demonstrations as in \cite{fewshot-cot}. Zero-Shot-CoT employs the template ``Let's think step by step''. We use GPT-3 with 175-billion parameters (text-davinci-002) as the LLM.
% Then we extend the generated CoT into the input of each sample in CSQA and StrategyQA and fine-tune ALBERT on the augmented datasets.
To perform fine-tuning on ALBERT, we extend the original input text by adding a CoT. We utilize ALBERT-large-v2 for our experiments. Specifically, we generate CoTs using both few-shot and zero-shot CoT methods, known as Few-Shot-CoT and Zero-Shot-CoT, respectively. 
Few-Shot-CoT employs the same demonstrations as described in \cite{fewshot-cot}. For Zero-Shot-CoT, we utilize the template ``Let's think step by step''. As the LLM, we employ GPT-3 with 175-billion parameters (text-davinci-002). Subsequently, we extend the generated CoT into the input of each sample within the CSQA and StrategyQA datasets. Finally, we perform fine-tuning on ALBERT using the augmented datasets.

% \begin{table}[]
% \centering
% \begin{tabular}{lcc}
% \toprule
% Method/Dataset & CSQA & StrategyQA \\ 
% \midrule
% \begin{tabular}[l]{@{}l@{}}Baseline\\ (ALBERT)\end{tabular}               & 63.4 & 64.8       \\ 
% \midrule
% \begin{tabular}[l]{@{}l@{}}Zero-Shot-CoT\\(ALBERT)\end{tabular} & 74.9 & 65.9       \\
% \begin{tabular}[l]{@{}l@{}}Few-Shot-CoT\\ (ALBERT)\end{tabular} & \textbf{76.2} & \textbf{73.1}       \\ 
% \bottomrule
% \end{tabular}
% % \caption{Accuracy (\%) of original fine-tuning (baseline) and CoT-augmented fine-tuning results. Zero-Shot-CoT and Few-Shot-CoT are two kinds of CoT generation methods.}
% \caption{Accuracy (\%) of original fine-tuning (baseline) and CoT-augmented fine-tuning results.}
% \label{table1-pilotstudy}
% \end{table}

% 0525
\begin{table}
\centering
% \resizebox{0.8 \linewidth}{!}{
\begin{tabular}{lcc}
\toprule
Method/Dataset & CSQA & StrategyQA \\ 
\midrule
\begin{tabular}[l]{@{}l@{}}Baseline\\ (ALBERT)\end{tabular}               & 63.4 & 64.8       \\ 
\midrule
\begin{tabular}[l]{@{}l@{}}Zero-Shot-CoT\\(ALBERT)\end{tabular} & 70.1 & 67.5       \\
\begin{tabular}[l]{@{}l@{}}Few-Shot-CoT\\ (ALBERT)\end{tabular} & \textbf{76.2} & \textbf{73.1}       \\ 
\bottomrule
\end{tabular}
% }
% \caption{Accuracy (\%) of original fine-tuning (baseline) and CoT-augmented fine-tuning results. Zero-Shot-CoT and Few-Shot-CoT are two kinds of CoT generation methods.}
\caption{Accuracy (\%) of original fine-tuning (baseline) and CoT-augmented fine-tuning results.}
\label{table1-pilotstudy}
\end{table}

The experiment results in Table \ref{table1-pilotstudy} show that both the Zero-Shot-CoT and Few-Shot-CoT augmented fine-tuning significantly enhance the performance of the original fine-tuning method.

\vpara{The Impact of CoT as Additional Knowledge.}
Given that the answers within CoTs can potentially be incorrect, we hypothesize that this portion of the CoTs will have a negative effect on the fine-tuning and mislead the model's prediction. To further explore the effect of CoTs on fine-tuning, we compare the fine-tuning result of the PLMs before and after adding CoTs through a variety of data analyses.
% Given that the answers within CoTs can potentially be incorrect, we put forth the hypothesis that this aspect of the CoTs could adversely impact the fine-tuning process and potentially lead the model astray in its predictions. In order to delve deeper into the impact of CoTs on fine-tuning, we conduct various data analyses to compare the results of fine-tuning PLMs both before and after adding CoTs.

% We examine how much the prediction results change when the model's input is expanded by a CoT. We fine-tune the original samples (baseline) and the expanded samples (CoT-extended), and evaluated the fine-tuned models on the validation set. 
% For each instance in the validation set, we compare its predictive result with the originally fine-tuned ALBERT and the CoT-augmented fine-tuning version and define three kinds of CoTs.

We investigate the extent to which the prediction results are altered when the model's input is expanded with a CoT. We perform fine-tuning on both the original samples (baseline) and the expanded samples (CoT-extended). Subsequently, we evaluate the fine-tuned models using the validation set. 
For each instance in the validation set, we compare its predictive result between the originally fine-tuned ALBERT and the CoT-augmented fine-tuning version. Additionally, we define three categories of CoTs during the process.

\begin{itemize}[leftmargin=*]
    % \item A CoT is regarded to have a positive impact on the model's prediction If the prediction result changes from incorrect to correct with the addition of a CoT, and this CoT is referred to as a \emph{positive CoT}. 
    \item A CoT is labeled as a \emph{positive CoT} if the addition of the CoT changes the prediction result from incorrect to correct. This indicates a beneficial influence on the model's prediction.
    % \item Conversely, we consider a CoT to have a misleading influence on the prediction of the model if the prediction result changes from correct to incorrect after adding a CoT. We label this CoT as a \emph{negative CoT}. 
    \item Conversely, a CoT is labeled as a \emph{negative CoT} if the addition of the CoT changes the prediction result from correct to incorrect. This indicates a misleading effect on the model's prediction.
    % \item In addition, we label a CoT \emph{neutral CoT} if the model's prediction result remains the same after adding a CoT. In this case, it is not easy to judge the impact of this CoT on the model. 
    \item Furthermore, a CoT is labeled as a \emph{neutral CoT} if the model's prediction result remains the same after the CoT is added. In such cases, it is not easy to judge the impact of this CoT on the model.
\end{itemize}

% \begin{figure}[t]
%     \centering
%     \includegraphics[width=\linewidth]{figures/figure2_3.pdf}
%     \caption{The observation when the original question added a CoT. The figure on the left shows the ratio of \emph{positive}, \emph{neutral}, and \emph{negative CoTs} in the validation set of StrategyQA. The figure on the right shows the proportion of model predictions that do not align with the answer in the CoT. ``Not misled'' denotes that the answer in CoT is incorrect, but the model is not misled by the CoT and makes accurate predictions. ``Not inspired'' denote that the answer in CoT is correct, but the model does not follow the correct CoT and makes the incorrect predictions.}
%     \label{fig23:observation}
% \end{figure}

\begin{figure}[t]
    \centering
    \includegraphics[width=\linewidth]{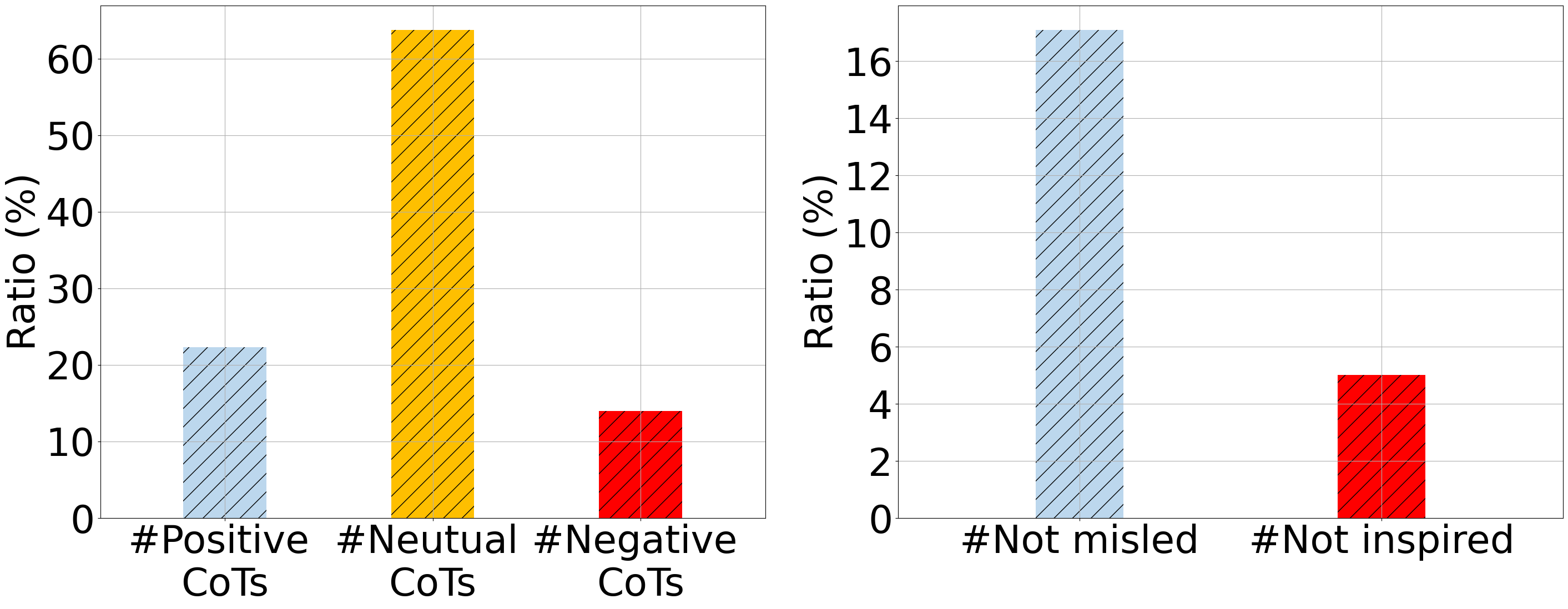}
    \caption{The observation when the original question added a CoT. The figure on the left shows the ratio of \emph{positive}, \emph{neutral}, and \emph{negative CoTs} in the validation set of StrategyQA. The figure on the right shows the proportion of model predictions that do not align with the answer in the CoT. ``Not misled'' denotes that the answer in the CoT is incorrect, but the model is not misled by the CoT and makes accurate predictions. ``Not inspired'' denotes that the answer in the CoT is correct, but the model does not follow the correct CoT and makes incorrect predictions.}
    \label{fig23:observation}
\end{figure}

% The left figure in Figure \ref{fig23:observation} displays the ratio of \emph{positive}, \emph{neutral}, and \emph{negative CoTs}. We observe that the ratio of the model predicted results that change after adding a CoT is 31.0\% (142/458), among which the ratio of positive CoT is 62.7\% and the ratio of negative CoT is 37.3\%. These results indicate that the model's resolution for the data samples that are incorrectly predicted before adding a CoT is 58.6\% (89/152, the number of positive CoTs / the number of incorrectly predicted samples in baseline).

The left figure in Figure \ref{fig23:observation} illustrates the ratio of \emph{positive}, \emph{neutral}, and \emph{negative CoTs}. 
% It is observed that among the model's prediction results that change after adding a CoT, the ratio is 31.0\% (142 out of 458). 
It is observed that among the model's prediction results that change after adding a CoT, the ratio is 36.2\% (166 out of 458). 
% 102 pos, 64 neg
% Within this group, the ratio of \emph{positive CoTs} is 61.7\%, while the ratio of \emph{negative CoTs} is 37.3\%. 
Within this group, the ratio of \emph{positive CoTs} is 61.4\%, while the ratio of \emph{negative CoTs} is 38.6\%. 
% These findings suggest that the model successfully resolves 58.6\% (89/152, the number of positive CoTs divided by the number of incorrectly predicted samples in the baseline) of the data samples that were incorrectly predicted prior to adding a CoT.
These findings suggest that the model successfully resolves 63.3\% (102/161, the number of positive CoTs divided by the number of incorrectly predicted samples in the baseline) of the data samples that were incorrectly predicted prior to adding a CoT.

The second objective is to test our hypothesis that an incorrect CoT (the answer in the CoT is incorrect) may have a negative impact on the model and therefore mislead the prediction of the model.
If an incorrect CoT is added to the original input text, what impact does it have on the model's prediction? 
% As the right figure in Figure \ref{fig23:observation} shows, when an incorrect CoT is added to the original input, the model still has a high probability (27.6\%) of not being misled by the incorrect CoT and making accurate predictions. Furthermore, we investigate the extent to which the model would mispredict when a correct CoT is added. As shown in the figure on the right of Figure \ref{fig23:observation}, the model has a low probability (7.5\%) of making an incorrect prediction.
As the right figure in Figure \ref{fig23:observation} shows, when an incorrect CoT is added to the original input, the model still has a high probability (17.1\%) of not being misled by the incorrect CoT and making accurate predictions. Furthermore, we investigate the extent to which the model would mispredict when a correct CoT (the answer in the CoT is correct) is added. As shown in the figure on the right of Figure \ref{fig23:observation}, the model has a low probability (5.0\%) of making an incorrect prediction.

In the case of StrategyQA, when the answer in the CoT is incorrect, the alignment ratio is \(1 - Ratio\ (\texttt{\#Not misled})\), which equals 82.9\%; When the answer in the CoT is correct, the alignment ratio is \(1 - Ratio\ (\texttt{\#Not inspired})\), which equals 95.0\%. The result 
% shown in the figure on the right of Figure \ref{fig23:observation} 
demonstrates that CoT is a powerful feature, and the model's predictions tend to align closely with the answers provided in CoT. 
% However, the fine-tuning strategy will cause the model's prediction to use the CoT as a secondary characteristic of the original input text rather than perfectly following it. When the CoT is correct, the model will follow the answer in CoT with a high probability. However, when the CoT is incorrect, there is also a good chance that the model will not keep up with the answer in the CoT, preventing misleading from the incorrect CoT.
On the other hand, the fine-tuning strategy employed causes the model's predictions to treat CoT as a secondary feature of the original input, rather than strictly following it. In cases where the answer in CoT is correct, the model is likely to align its predictions with the answers in CoT. Conversely, when the answer in CoT is incorrect, there is a relatively high probability that the model will deviate from the answer in the CoT, preventing misleading from the incorrect CoT.

In addition, our attempts to preserve the reasoning steps in the CoTs while removing the answers have resulted in a degradation in performance. We recognize that the presence of incorrect answers in some CoTs can have a negative impact. However, we also believe that the inclusion of correct answers in CoTs can yield positive effects, and the answers within CoTs are a more influential factor than the reasoning paths themselves.

\section{\model}
\begin{figure*}[ht]
    \centering
    \includegraphics[width=1.0\linewidth]{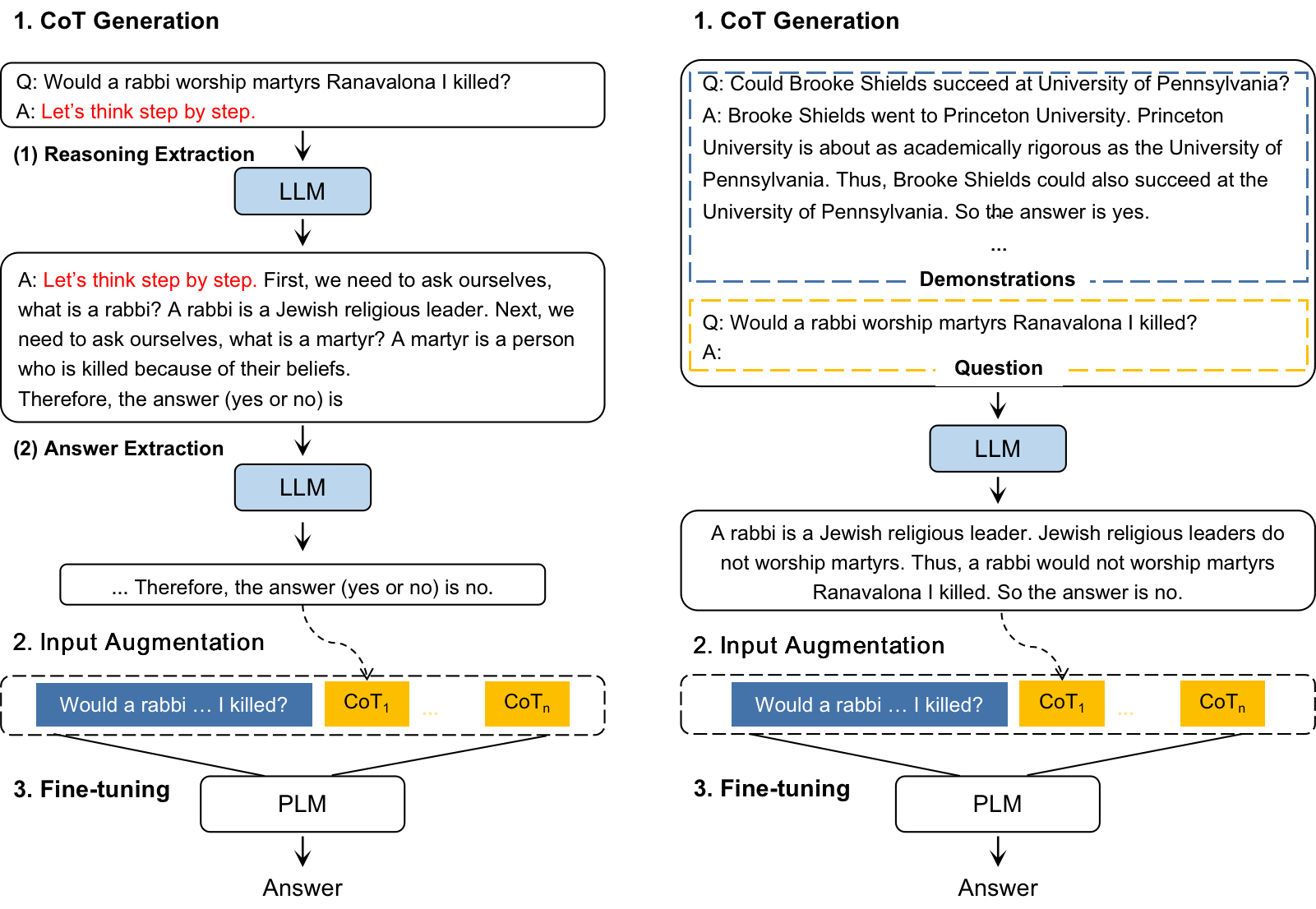}
    \caption{Overview of the CoT-KA method. Both Zero-Shot-CoT (on the left) and Few-Shot-CoT (on the right) can be used in the CoT generation stage for CoT-KA.}
    \label{fig4:CoT-KA}
\end{figure*}

In this section, we propose \smodel-- a CoT-based method for knowledge augmentation. Our method leverages multiple CoTs retrieved from LLMs to provide more auxiliary knowledge for KADL.
CoT-KA consists of three steps as shown in Figure \ref{fig4:CoT-KA}: 
% (1) CoT Generation: Generating multiple CoTs for each sample in the training, validation, and test sets. 
(1) CoT Generation: Generating multiple CoTs for each sample in the train, dev, and test sets. 
(2) Input Augmentation: Taking the generated CoTs as the additional knowledge into the original input text for each sample. 
(3) Task-relevant Model Training: Fine-tuning a task-relevant model using the CoT-augmented samples.

\subsection{CoT Generation}
% We take the CoTs generated from LLM as the additional knowledge for original input for each sample, 
We try both Few-Shot-CoT and Zero-Shot-CoT prompting on LLM \emph{f} to generate multiple CoTs. Formally, given an original samples \((x_i, y_i)\), where \(x_i\) is the original input and \(y_i \in \mathcal{Y}\) denotes the label. We generate a CoT set consisting of multiple CoTs based on the model \emph{f}:

\begin{align}
    CoT^{(i)} &= f(d, x_i)  \label{f}
\end{align}

\noindent where \emph{d} denotes the CoT demonstrations that inspire model \emph{f} to generate CoTs, 
and \(CoT^{(i)}\) is the generated CoT set of the $i$-th sample, which consists of $m$ CoTs:
\begin{align}
    CoT^{(i)} &= \{CoT_1^{(i)}, CoT_2^{(i)}, ..., CoT_m^{(i)}\} \label{cot}
\end{align}

For each sample, we independently generate $m$ CoT outputs from \emph{f} in each run. 

\subsection{Input Augmentation}
In the second step, we apply the generated CoTs as additional knowledge to enrich the input text of the original samples.
The extended input text of each sample is a concatenation of an original input (e.g. a question), and the generated multiple CoTs. For each sample, we construct an extended input text as follows: 

\begin{align}
    \tilde x^{(i)} = concat(x^{(i)}, CoT^{(i)}) \label{concat}
\end{align}

\noindent where $\tilde x^{(i)}$ is the \(i\)-th extended input text, 
$x^{(i)}$ is the \(i\)-th original input, and 
$CoT^{(i)}$ is the  \(i\)-th generated CoT set.
$concat()$ is a concatenation function that concatenates the original input and the generated CoTs. More concretely: 

\begin{align}
    & concat(x^{(i)}, CoT^{(i)}) \notag \\ 
    &= x^{(i)} || \ [EXT] \ CoT_1^{(i)} ...\ || \ [EXT] \  CoT_m^{(i)} 
\end{align}

% \verb|[EXT]|
\noindent where %\( x^{(i)} || \ [EXT] \ CoT_1^{(i)} ...\ || \ [EXT] \  CoT_m^{(i)} \) is the input sequence to be encoded into contextual token embeddings with pre-trained language models. 
\([EXT]\) is the special token to denote a CoT, and \(||\) denotes the concatenation operator.

% \begin{align}
%     & concat(x^{(i)}, CoT^{(i)}) \notag \\ 
%     &= x^{(i)} || \ \texttt{[EXT]} \ CoT_1^{(i)} ...\ || \ \texttt{[EXT]} \  CoT_m^{(i)} 
% \end{align}

% \noindent where \texttt{[EXT]} is the special token to denote a CoT, and \(||\) denotes the concatenation operator.

\section{Experiments}
\subsection{Experimental setup}

\paragraph{Tasks and Datasets.}
We evaluate \smodel on the following reasoning benchmarks\footnote{By default we use the train, dev, and test split of all the datasets if the labels are available for evaluation. For CSQA and StrategyQA, we only use the train and dev split.}.

\begin{table*}[t]
\centering
\resizebox{\linewidth}{!}{
\begin{tabular}{lcccccccc}
\toprule
\multicolumn{1}{c}{}                                                                & \multicolumn{6}{c}{\textit{Commonsense}}                                                                      & \multicolumn{2}{c}{\textit{Arithmetic}} \\ 
\hline
\multicolumn{1}{l}{\multirow{2}{*}{Method/Dataset}}                                 & CSQA          & StrategyQA    & \multicolumn{2}{c}{Date}              & \multicolumn{2}{c}{Sports}            & \multicolumn{2}{c}{AQuA}                \\
\multicolumn{1}{c}{}                                                                & Dev           & Dev           & Dev               & Test              & Dev               & Test              & Dev                & Test               \\ \hline
Zero-Shot-CoT                                                                       & 64.6*         & 54.8*         & \multicolumn{2}{c}{{\ul 67.5}*}             & \multicolumn{2}{c}{52.4*}             & \multicolumn{2}{c}{33.5*}               \\
Few-Shot-CoT                                                                        & - (73.5*)       & 68.3 (65.4*)  & \multicolumn{2}{c}{54.7/47.4 (52.1*)} & \multicolumn{2}{c}{83.2/86.7 (82.4*)} & \multicolumn{2}{c}{-/37.9 (35.8*)}      \\
\begin{tabular}[c]{@{}l@{}}Self-Consistency\\ (5 Zero-Shot-CoTs)\end{tabular} & 71.2          & 64.6          & \multicolumn{2}{c}{29.2/35.6}   & \multicolumn{2}{c}{57.6/58.9}         & \multicolumn{2}{c}{33.2/37.0}           \\
\begin{tabular}[c]{@{}l@{}}Self-Consistency\\ (5 Few-Shot-CoTs)\end{tabular}  & 77.6          & 73.6          & \multicolumn{2}{c}{53.4/50.1}         & \multicolumn{2}{c}{85.4/90.5}         & \multicolumn{2}{c}{40.6/40.2}           \\ \hline 
\begin{tabular}[c]{@{}l@{}}Baseline\\ (ALBERT)\end{tabular}                         & 61.8          & 62.2          & 33.2              & 33.5              & 57.2              & 53.2              & 25.6               & 22.7               \\
\begin{tabular}[c]{@{}l@{}}CoT-KA \\ (5 Zero-Shot-CoTs, ALBERT)\end{tabular}        & 73.6          & 66.1          & 58.6              & 64.1              & 68.8              & 69.6              & 42.3               & 40.2               \\
\begin{tabular}[c]{@{}l@{}}CoT-KA \\ (5 Few-Shot-CoTs, ALBERT)\end{tabular}         & \textbf{78.8} & \textbf{75.7} & \textbf{74.2}     & \textbf{76.6}     & \textbf{89.9}     & \textbf{89.8}     & \textbf{46.9}      & \textbf{47.6}      \\ \hline
\begin{tabular}[c]{@{}l@{}}Baseline\\ (DeBERTa)\end{tabular}                        & \textbf{84.2} & 68.8          & 73.6              & 72.7              & 84.5              & 82.8              & 27.8               & 26.5               \\
\begin{tabular}[c]{@{}l@{}}CoT-KA \\ (5 Zero-Shot-CoTs, DeBERTa)\end{tabular}       & 80.3          & 72.3          & 69.2              & 73.8              & 91.3              & 90.5              & 40.1               & 40.3               \\
\begin{tabular}[c]{@{}l@{}}CoT-KA \\ (5 Few-Shot-CoTs, DeBERTa)\end{tabular}        & 82.0          & \textbf{76.9} & \textbf{80.4}     & \textbf{78.0}     & \textbf{96.9}     & \textbf{95.6}     & \textbf{45.9}      & \textbf{46.5}      \\ 
\bottomrule
\end{tabular}
}
\caption{Accuracy on five NLU datasets from two categories of reasoning tasks. For CSQA and StrategyQA, we report the evaluation results of the dev set. For the other datasets in which the labels are available, we report the results of both the dev and test. * indicates the results comes from \cite{fewshot-cot} and \cite{zeroshot-cot}. The results of baseline methods and CoT-KA are based on ALBERT-large-v2 and DeBERTa-v3-large. ``Baseline'' denotes the fine-tuning baseline with original data. ``5 Zero-Shot-CoTs'' and ``5 Few-Shot-CoTs'' denotes five CoTs used at Self-Consistency and CoT-KA. Bold denotes the best-performed results.
For Few-Shot-CoT, the results before and after the ``/'' symbol indicate the results of directly parsing the answers from the CoT (from \citet{fewshot-cot}) for the dev and test set, respectively, under our data partitioning.
For Self-Consistency, the results before and after the ``/'' symbol represent the results obtained by parsing the answer from multiple CoTs (We generated) in the dev and test set, respectively, under our data partitioning and then applying majority voting.
}
\label{tab:main1}
\end{table*}

\begin{table*}[t]
\centering
\resizebox{\linewidth}{!}{
% \usepackage{multirow}
\begin{tabular}{lcccccccccccc}
\toprule
\multicolumn{1}{c}{}                                                          & \multicolumn{10}{c}{\textit{Arithmetic}}                                                                                                                                                           & \multicolumn{2}{c}{\textit{Symbolic}} \\ \hline
\multicolumn{1}{l}{\multirow{2}{*}{Method/Dataset}}                           & \multicolumn{2}{c}{GSM8K}          & \multicolumn{2}{c}{SVAMP}             & \multicolumn{2}{c}{MultiArith}        & \multicolumn{2}{c}{SingleEq}          & \multicolumn{2}{c}{AddSub}            & \multicolumn{2}{c}{Letter (4)}        \\
\multicolumn{1}{c}{}                                                          & Dev              & Test            & Dev           & Test                  & Dev                & Test             & Dev                & Test             & Dev               & Test              & Dev               & Test              \\ \hline
Zero-Shot-CoT                                                                 & \multicolumn{2}{c}{40.7*}          & \multicolumn{2}{c}{63.7*}             & \multicolumn{2}{c}{78.7*}             & \multicolumn{2}{c}{78.7*}             & \multicolumn{2}{c}{74.7*}             & \multicolumn{2}{c}{57.6*}             \\
Few-Shot-CoT                                                                  & \multicolumn{2}{c}{-/46.5 (46.9*)} & \multicolumn{2}{c}{69.2/69.0 (68.9*)} & \multicolumn{2}{c}{85.8/90.0 (91.7*)} & \multicolumn{2}{c}{82.4/87.3 (86.6*)} & \multicolumn{2}{c}{79.7/65.8 (81.3*)} & \multicolumn{2}{c}{(59.0**)}          \\
\begin{tabular}[c]{@{}l@{}}Self-Consistency\\ (5 Zero-Shot-CoTs)\end{tabular} & \multicolumn{2}{c}{51.7/52.2}      & \multicolumn{2}{c}{70.0/73.4}         & \multicolumn{2}{c}{81.7/\textbf{96.4}}         & \multicolumn{2}{c}{64.8/92.0}         & \multicolumn{2}{c}{79.7/73.7}         & \multicolumn{2}{c}{66.3/60.2}         \\
\begin{tabular}[c]{@{}l@{}}Self-Consistency\\ (5 Few-Shot-CoTs)\end{tabular}  & \multicolumn{2}{c}{55.7/56.6}      & \multicolumn{2}{c}{\textbf{74.7}/75.5}         & \multicolumn{2}{c}{\textbf{94.8}/95.7}         & \multicolumn{2}{c}{\textbf{88.5}/\textbf{91.9}}         & \multicolumn{2}{c}{86.8/73.9}         & \multicolumn{2}{c}{59.0/60.5}         \\ \hline
Baseline (T5)                                                                 & 5.3              & 4.4             & 8.0           & 8.5                   & 12.5               & 8.3              & 5.9                & 2.9              & 6.3               & 6.3               & 30.0              & 26.0              \\
\begin{tabular}[c]{@{}l@{}}CoT-KA\\ (5 Zero-Shot-CoTs, T5)\end{tabular}       & 58.9             & 57.3            & 64.2          & \textbf{82.3}         & 82.7               & 93.3             & 62.9               & 73.3             & 80.3              & 74.9              & \textbf{75.9}     & 60.4              \\
\begin{tabular}[c]{@{}l@{}}CoT-KA\\ (5 Few-Shot-CoTs, T5)\end{tabular}        & \textbf{61.2}    & \textbf{61.5}   & 71.8          & 70.8                  & 81.8               & 95.3             & 76.7               & 75.7             & \textbf{86.6}     & \textbf{78.7}     & \textbf{71.8}              & \textbf{69.8}     \\ 
\bottomrule
\end{tabular}
}
\caption{Accuracy on six NLG datasets from two categories of reasoning tasks. * indicates the results comes from \cite{fewshot-cot} and \cite{zeroshot-cot} and ** denotes the result comes from \cite{auto-cot}.}
\label{tab:main2}
\end{table*}

\begin{itemize}[leftmargin=*]
    \item \textbf{Commonsense reasoning.} We evaluate our method on four commonsense reasoning tasks: CSQA ~\cite{CSQA}, StrategyQA ~\cite{strategyQA} and two benchmarks from the BIG-bench effort ~\cite{big-bench}: Date Understanding and Sports Understanding.
    \item \textbf{Arithmetic reasoning.} We use six arithmetic reasoning benchmarks: AQUA-RAT \cite{aqua}, GSM8K \cite{gsm8k}, SVAMP \cite{svamp}, MultiArith \cite{multiarith}, SingleEq \cite{singleeq}, AddSub \cite{addsub}.
    \item \textbf{Symbolic Reasoning.} We use the Last Letter Concatenation from \citet{fewshot-cot}. \footnote{We do not use the Coin Flip dataset for the evaluation because it is a simple classification task for fine-tuning. This is because ALBERT-large-v2 and DeBERTa-v3-large can already achieve 100\% accuracy in the evaluation phase.}
\end{itemize}

\begin{table*}[ht]
\centering
\resizebox{1.0\textwidth}{!}{
\begin{tabular}{ll}
\toprule
StrategyQA & \begin{tabular}[c]{@{}l@{}}\textbf{Question}: Would Siduri enjoy an unlimited buffet?\\ \textcolor{color1}{\textbf{Blink}}: Siduri is a character in the "Epic of Gilgamesh". She is an "alewife",  a wise female \\ divinity associated with fermentation (specifically beer and wine).\\ \textcolor{color3}{\textbf{Few Shot CoT}}: Siduri is a fairy in Irish mythology. She was known for her hospitality, so \\ she would probably enjoy an unlimited buffet. So the answer is yes.\end{tabular}                                                                                  \\ \hline
Sports     & \begin{tabular}[c]{@{}l@{}}\textbf{Question}: Will Fuller was perfect from the line?\\ \textcolor{color1}{\textbf{Blink}}: William Vincent Fuller V (born April 16, 1994) is an American football wide receiver \\ for the Houston Texans of the National Football League (NFL). He was drafted by the Texans \\ in the first round of the 2016 NFL Draft. He played college football at Notre Dame.\\ \textcolor{color3}{\textbf{Few Shot CoT}}: Will Fuller is a football player. Being perfect from the line is part of basketball, \\ not football. So the answer is no.\end{tabular} \\ \bottomrule
\end{tabular}
}
\caption{Knowledge augmentation examples from commonsense reasoning tasks. The first case comes from StrategyQA. In this case, the description of Siduri does not mention the relationship between Siduri and the unlimited buffet, which is the key to answering the question. 
The second case comes from Sports Understanding. In this case, we need to know that being perfect from the line is part of basketball, and Will Fuller is a football player, while the entity-knowledge can only provide the latter.}
\label{tab:knowledge-aug-cases}
\end{table*}

\paragraph{Implementation.}
\begin{itemize}[leftmargin=*]
    % \item \textbf{CoT Generation Models.} We use GPT-3 of with 175-billion parameters to generate the CoTs used in CoT-KA. We use the text-davinci-002 engine used in Few-Shot-CoT and Zero-Shot-CoT.
    \item \textbf{CoT Generation Models.} We use GPT-3 of the text-davinci-002 engine with 175-billion parameters to generate the CoTs used in CoT-KA. 
    \item \textbf{CoT Demonstrations.} For a fair comparison, we perform Few-Shot-CoT with the same demonstrations as in \citet{fewshot-cot} and use the same zero-shot prompt as in \citet{zeroshot-cot} to perform Zero-Shot-CoT.
    \item \textbf{Sampling Scheme.} To generate diverse CoTs, we apply temperature sampling during the CoT generation. Specifically, we use the same \emph{T}=0.7 as in \cite{self-consistency} for a fair comparison.
    \item \textbf{Data Preprocessing.} For certain undivided datasets, we divide them into train, dev, and test sets for fine-tuning, following a ratio of 6:2:2. Further details regarding the dataset splits can be found in Appendix \ref{appendix:datasets}. Additionally, as the original questions and demonstrations used for CoT generation may include option information (e.g., Answer Choices: \emph{(a) ignore ...(e) avoid}), the generated CoT will also contain option markers (e.g., the answer is \emph{(a)}). To provide valuable information within the CoTs, we replace the option markers in the generated CoT with their corresponding textual content (e.g., the answer is \emph{``ignore''}).
    \item \textbf{Classifier Models.} We conduct the main experiments using two backbone PLMs: ALBERT-large-v2 and DeBERTa-v3-large. The hyper-parameters for the training process are reported in Appendix \ref{appendix:implementation}.
\end{itemize}

\paragraph{Baselines.}
We take three methods as the baselines: Zero-Shot-CoT, Few-Shot-CoT, and Self-Consistency. Furthermore, to demonstrate the extent to which the CoT knowledge elicits the KADL, 
we also compare our method with the original fine-tuning baselines, which solely employ the original text for fine-tuning.

\subsection{Main Results}
Table \ref{tab:main1} compares the accuracy across eleven datasets from three categories of NLU and NLG tasks. 
The Zero-Shot-CoT results are taken from \citet{zeroshot-cot}, and the Few-Shot-CoT results are taken from \citet{fewshot-cot}. 
% We also use a majority vote to report the result of Self-Consistency (5 sampled CoTs). 
For Self-Consistency (5 sampled CoTs), we report the result based on a majority vote.
The \smodel results are averaged over at least five random runs (see Appendix for more details), where we use the different seeds to sample 5 CoTs from a CoT set containing 10 generated CoTs in each run. 

As shown in Table \ref{tab:main1} and \ref{tab:main2}, the performance of \smodel surpasses all baselines on most tasks. We have made several findings:
(1) The CoTs generated by Zero-Shot-CoT and Few Shot-CoT can be utilized with \model, resulting in significantly improved performance compared to the fine-tuning baselines.
% This shows that adding CoTs generated by LLMs to the original text can be effective for KADL.
% Additionally, the CoTs generated by Few-Shot-CoT have obtained better performance than Zero-Shot-CoT in \model.
Additionally, the CoTs generated by Few-Shot-CoT exhibit better performance compared to Zero-Shot-CoT when they are used with \model. 
(2) \smodel achieves better performance on the NLU tasks than on the NLG tasks.
    % \item  However, this superior performance is not significant on xx tasks. This indicates that the performance of Zero-Shot-CoT Extension can approximate that of Few-Shot-CoT Extension in the scenario of extending multiple CoTs.
(3) \smodel shows different robustness on different models. While DeBERTa outperforms ALBERT on most tasks, \smodel is more robust on ALBERT and exhibits performance improvements across all tasks.

\subsection{Knowledge Augmentation Comparison}

To compare \smodel with other knowledge augmentation methods, we employ BLINK \cite{BLink-} to enrich the entity knowledge in the question. BLINK is a two-stage entity linking approach based on BERT \cite{BERT}. We use BLINK to link the entities mentioned in the question and retrieve their corresponding entity information. BLINK provides a short description for each entity, which we utilize as extensions to enrich the questions. 

\begin{table}[ht]
\centering
\resizebox{1.0\linewidth}{!}{
\begin{tabular}{lclcc}
\toprule
\multicolumn{1}{l}{\multirow{2}{*}{Method/Dataset}} & \multicolumn{2}{c}{StrategyQA}    & \multicolumn{2}{c}{Sports} \\
\multicolumn{1}{c}{}                                & \multicolumn{2}{c}{Dev}           & Dev     & Test             \\ \hline
Baseline (ALBERT)                                   & \multicolumn{2}{c}{62.2}          & 57.2    & 53.2             \\
BLink (ALBERT)                                      & \multicolumn{2}{c}{58.0}          & 81.3    & 77.4             \\
CoT-KA (ALBERT)                                     & \multicolumn{2}{c}{75.7}          & 89.9    & 89.8             \\ \hline
Baseline (DeBERTa)                                  & \multicolumn{2}{c}{68.8}          & 84.5    & 82.8             \\
BLink (DeBERTa)                                     & \multicolumn{2}{c}{67.7}          & 92.5    & 87.5             \\
CoT-KA (DeBERTa)                                    & \multicolumn{2}{c}{\textbf{76.9}} &\textbf{96.9}    & \textbf{95.6}    \\ 
\bottomrule
\end{tabular}
}
\caption{Knowledge augmentation comparison.}
\label{tab:Blink}
\end{table}

As shown in Table \ref{tab:Blink}, 
% the augmentation method based on entity knowledge improves the performance in Sports Understanding dataset but does harm  StrategyQA, and both of them are worse than our method. 
the entity knowledge-based augmentation method improves performance on Sports Understanding but has a negative impact on StrategyQA, with both performing worse than our method. 
% It is shown that we cannot extract entities from about 29\% questions in StrategyQA and 3\% in Sports Understanding. We also find that the average number of entities recognized in a question in Sports Understanding is 1.095, and in StrategyQA, this number is 0.928. Besides, as Table \ref{tab:knowledge-aug-cases} shows, the entity information may not include the specific information needed by the questions. However, our method can add more useful information, which leads to larger improvement.
Additionally, we observe that approximately 29\% of questions in StrategyQA and 3\% in Sports Understanding could not have entities extracted. Furthermore, the average number of recognized entities in a Sports Understanding question is 1.095, while in StrategyQA, it is 0.928. Moreover, Table \ref{tab:knowledge-aug-cases} demonstrates that entity information may not always include the specific information required by the questions. In contrast, our method can add more useful information, resulting in a more substantial improvement.

\subsection{The Effect of CoT Size}
\label{sec-cot-size}
% \begin{figure*}[htp]
%     \centering
%     \includegraphics[width=\textwidth]{figures/figure5_cot-nums.pdf}
%     \caption{The impact of the sampled CoT size on \smodel. We randomly sampled 1 to 5 CoTs from both the CoT set generated by Zero-Shot-CoT and Few-Shot-CoT.}
%     \label{fig:cotnums}
% \end{figure*}
\begin{figure}[t]
    \centering
    \includegraphics[width=\linewidth]{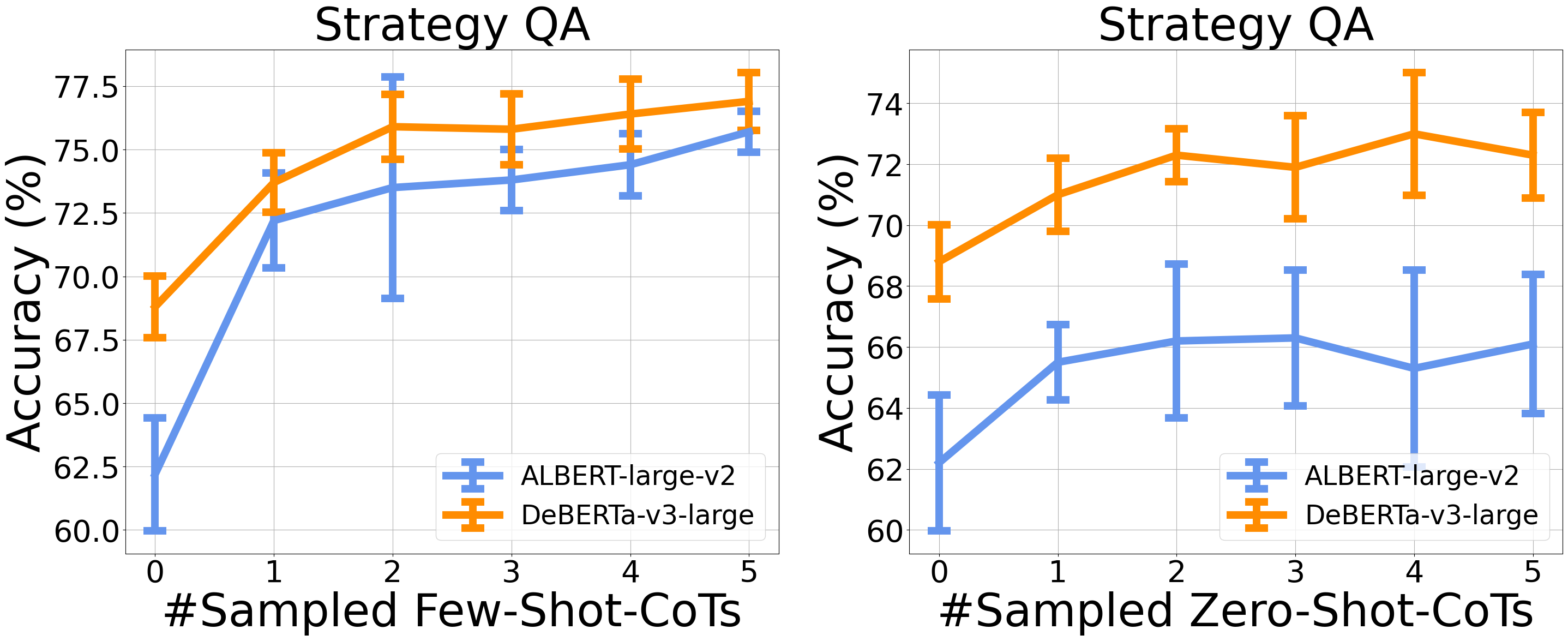}
    \caption{The impact of the sampled CoT size on \smodel. We randomly sampled 1 to 5 CoTs from both the CoT set generated by Zero-Shot-CoT and Few-Shot-CoT.}
    \label{fig:cotnums}
\end{figure}

To demonstrate the effect of the number of sampled CoTs, we vary the number of sampled CoTs (1, 2, 3, 4, 5) in \smodel and evaluate on StrategyQA. The results are shown in Figure \ref{fig:cotnums}. 
The experimental results indicate that as the number of CoTs increases, 
% the performance of \smodel generally presents an upward trend. 
there is a general upward trend in the performance of \smodel. 
% If the CoT is generated by Few-Shot-CoT, this tendency will be more evident. 
This trend becomes more pronounced when the CoTs are generated by Few-Shot-CoT.
More results are reported in Appendix \ref{appdix: the Effect}.

\subsection{CoT Selection Strategy}

\smodel can only extend a small number of CoTs due to the maximum length limitation of the input sequence that the language model can handle. 
% Therefore, the natural idea is to design a CoT selection strategy for selecting higher-quality CoTs from the generated CoT set for KADL. 
Therefore, it is natural to consider designing a CoT selection strategy to choose higher-quality CoTs from the generated CoT set for KADL.
Each CoT can be expressed as:
\(
t_i\in \{t_{1},t_{2}, ..., t_{K}\}
\)
, where $t_i$ is the \emph{i}-th token.
% \emph{logprob}
We can get the \emph{log prob} of each generated token when using GPT3 API to generate reasoning chains. The \emph{log prob} refers to the natural logarithm of the probability that the token occurs next given the prompt. To select the 5 reasoning chains with higher confidence from the 10 generated CoTs, we score the generated CoTs using the following formula:
% \begin{align}
%     Score(r_{i}) = \frac{\sum_{i=1}^{K}\text{exp}(\text{log P}(t_i))}{K} 
%     = \frac{\sum_{i=1}^{K}\text{P}(t_i)}{K} \nonumber
% \end{align}
% \begin{align}
%     score(CoT_j) = \frac{\sum_{i=1}^{K_j}\exp(\text{log } p(t_i))}{K_j} 
%     = \frac{\sum_{i=1}^{K_j}p(t_i)}{K_j} \nonumber
% \end{align}
\begin{align}
    score(CoT_j) &= \frac{\sum_{i=1}^{K_j}\exp(\text{log } p(t_i))}{K_j} \notag \\ 
    &= \frac{\sum_{i=1}^{K_j}p(t_i)}{K_j}
\end{align}
% \begin{align}
%     & concat(x^{(i)}, CoT^{(i)}) \notag \\ 
%     &= x^{(i)} || \ [EXT] \ CoT_1^{(i)} ...\ || \ [EXT] \  CoT_m^{(i)} 
% \end{align}
% where \(\text{log } p(t_i)\) denotes the logarithm of the probability of generating the \(i\)-th token, \(p(t_i)\) denotes the probability of generating the \(i\)-th token, 
where \(p(t_i)\) denotes the probability of generating the \(i\)-th token, and log denotes the logarithm.
% and \emph{K} is the total number of tokens in a reasoning chain \(r_i\).
and \(K_j\) is the total number of tokens in the \(j\)-th CoT.
The results shown in Table \ref{tab:cot selection} demonstrate that selecting CoTs from the generated set based on the probability of token generation in the sentence does not lead to a significant improvement in the performance of \smodel.
\begin{table}[ht]
\centering
\resizebox{1.0\linewidth}{!}{
% \begin{tabular}{lc}
% \toprule
% \multicolumn{1}{c}{\multirow{2}{*}{Method}} & \multirow{2}{*}{StrategyQA} \\
% \multicolumn{1}{c}{}                        &                             \\ \hline
% CoT-KA (ALBERT)                             & 75.7                        \\
% CoT-KA (ALBERT) + CoT Selection             & 75.9                        \\ \hline
% CoT-KA (DeBERTa)                            & 76.9                        \\
% CoT-KA (DeBERTa) + CoT Selection            & 76.9                        \\ 
% \bottomrule
% \end{tabular}

\begin{tabular}{lc}
\toprule
Method           & StrategyQA \\ \hline
CoT-KA (ALBERT)  & 75.7           \\
CoT-KA (ALBERT) + CoT Selection  & 75.9           \\ \hline
CoT-KA (DeBERTa) & 76.9           \\
CoT-KA (DeBERTa) + CoT Selection & 76.9           \\ \bottomrule
\end{tabular}
}
\caption{CoT selection strategy based on the \emph{log prob}}
\label{tab:cot selection}
\end{table}

\section{Conclusion and Future Work}
\label{sec:conclusion}
This paper introduces a \underline{CoT}-based method to retrieve knowledge from LLMs for \underline{K}nowledge-\underline{A}ugmented deep learning (\model) that elicits knowledge augmentation on a variety of NLU and NLG benchmarks.  
% Our method does not needs a retriever or a reasoner to perform knowledge augmentation but still exceeds the performance of the conventional knowledge and other CoT-based methods at a variety of public NLP tasks.
Unlike conventional knowledge augmentation approaches, our method does not require a retriever or a reasoner, yet it surpasses the performance of conventional knowledge-based methods and other CoT-based approaches across a range of public NLP tasks.

In the future, it is worthwhile to investigate other methods that can provide insights from LLMs. 
Exploring new approaches for leveraging the capabilities of LLMs to enhance knowledge augmentation represents a promising area for future research.

\section{Limitations}
One limitation of \smodel is that it performs fine-tuning based on the PLMs, and the input sequence length limit of the PLMs allows us to add only a limited number of CoTs. 
% So it is worth investigating how to design a CoT selection strategy in the future. 
Therefore, it is important to explore and develop a CoT selection strategy in future research. 
% A good CoT selection strategy will guide us to choose some CoTs from among multiple CoTs that are more efficient for KADL. 
A good CoT selection strategy would enable the identification of highly effective CoTs from a set of CoTs, enhancing the efficiency of KADL. 
% Additionally, the use of new methods instead of CoT Prompting to gain LLM insight is also an area worthy of further research.

\section*{Acknowledgments}
This work is supported by National Natural Science Foundation of China 62076245; CCF-Zhipu AI Large Model Fund.

\bibliography{custom}
\bibliographystyle{acl_natbib}

\newpage
\appendix
\section{Implementation Detail}
\label{sec:appendix}

\subsection{Datasets}
\label{appendix:datasets}

\begin{table}[ht]
\centering
\resizebox{1.0 \linewidth}{!}{
\begin{tabular}{lcccc}
% \begin{tabular}{m<{\centering}m<{\centering}m<{\centering}m<{\centering}m<{\centering}}
% \hline
\toprule
\multirow{2}{*}{Dataset}  & \multicolumn{3}{c}{\#Number of samples} & \multirow{2}{*}{We divide the dataset} \\ \cline{2-4}
            & Train       & Dev        & Test       &                                 \\ \midrule
CSQA        & 9741        & 1221       & 1140       & No                              \\
StrategyQA  & 1831        & 458        & 490        & No                              \\
Date        & 221         & 74         & 74         & Yes                             \\
Sports      & 600         & 200        & 200        & Yes                             \\
AQUA        & 5000        & 254        & 254        & Yes                           \\
GSM8K       & 5978        & 1495       & 1319       & Yes                             \\
SVAMP       & 600         & 200        & 200        & Yes                             \\
MuitiArith  & 360         & 120        & 120        & Yes                             \\
Single Eq   & 304         & 102        & 102        & Yes                             \\
Add Sub     & 237         & 79         & 79         & Yes                             \\
Last Letter & 600         & 200        & 200        & Yes                             \\ \bottomrule
\end{tabular}
}
\caption{Summary of the datasets we use in this paper. For datasets that are not pre-divided into train, dev, and test sets, we conduct the division ourselves.}
\label{tab:dataset-split-detail}
\end{table}

For some undivided datasets used in this paper, we divide them into train, dev, and test sets for fine-tuning, following a ratio of 6:2:2. Table \ref{tab:dataset-split-detail} shows the division details of each dataset. 
% The raw training set for AQUA is too large (97467 samples). To avoid the excessive computational cost caused by using the public GPT3 API to generate multiple CoTs, we only selected 5000 data from the raw training set as our training set.
In the case of AQUA, the raw training set is too large (97467 samples). To mitigate the computational cost of generating multiple CoTs using the public GPT3 API, we select a subset of 5000 samples (the top 5000) from the raw train set as our train set.

\subsection{Hyper-parameters for Fine-tuning}
\label{appendix:implementation}
All experiments are conducted in a Linux environment with a single (24G) NVidia RTX 3090 GPU. The model is optimized using the AdamW optimizer.
We do not perform an exhaustive hyper-parameter search, but only adjust the learning rate prior to the formal experiment. For most experiments in this paper, a learning rate of 1e-5 is chosen as the final value for fine-tuning ALBERT and DeBERTa, except in the following cases for CSQA and StrategyQA: 
\begin{itemize}
    \item CSQA: A learning rate of 2e-5 is used for CoT-KA (1 Zero-Shot-CoT, ALBERT).
    \item StrategyQA: A learning rate of 5e-6 is used for CoT-KA (1 Zero-Shot-CoT, ALBERT), CoT-KA (1 Few-Shot-CoT, DeBERTa) and CoT-KA (5 Few-Shot-CoTs, both ALBERT and DeBERTa).
\end{itemize}

More hyper-parameters are shown in Table \ref{tab:hyper-parameters}.

\begin{table}[ht]
\centering
\resizebox{0.9\linewidth}{!}{
\begin{tabular}{lcc}
% \hline
\toprule
                   & ALBERT/DeBERTa & T5    \\ % \hline
\midrule
Batch Size         & 16             & 16    \\
Peak Learning Rate & 1e-5           & 1e-5  \\
Training Steps     & 2000           & 2000  \\
Warmup Proportion  & 0.1            & 0     \\
Weight Decay       & 0              & 0     \\
Adam \(\epsilon\)               & 1e-8           & 1e-8  \\
Adam \(\beta_{1}\)             & 0.9            & 0.9   \\
Adam \(\beta_{2}\)              & 0.999          & 0.999 \\ 
% \hline
\bottomrule
\end{tabular}
}
\caption{Hyper-parameters for fine-tuning.}
\label{tab:hyper-parameters}
\end{table}

The random seed set utilized for experiments is \texttt{[0, 10, 20, 30, 40, 50, 60, 70, 80, 90]}. These seeds are used for both CoT sampling and fine-tuning. For the case of experimental results averaged over five runs, we use the top five seeds from the seed set. For NLU tasks, most experimental results in Table \ref{tab:main1} are averaged over ten runs, except for the following cases:
\begin{itemize}
    \item CoT-KA (5 Zero-Shot-CoTs) on all NLU tasks are averaged over five runs. 
    \item CoT-KA (5 Few-Shot-CoTs) on AQUA is averaged over five runs. 
\end{itemize}

For NLG tasks, most results in Table \ref{tab:main2} are averaged over ten runs, with the exception of CoT-KA (5 Zero-Shot-CoTs) and CoT-KA (5 Few-Shot-CoTs), which are averaged over five runs. 

The result for Blink in Table \ref{tab:Blink} are averaged over five runs. All the new results in Section \ref{sec-cot-size} and Appendix \ref{appdix: the Effect}, where the number of sampled CoTs ranges from 1 to 4, are averaged over five runs.

\section{More results about the Effect of CoT Size in CoT-KA}
\label{appdix: the Effect}

\begin{figure*}[htb]
    \centering
    \includegraphics[width=1 \linewidth]{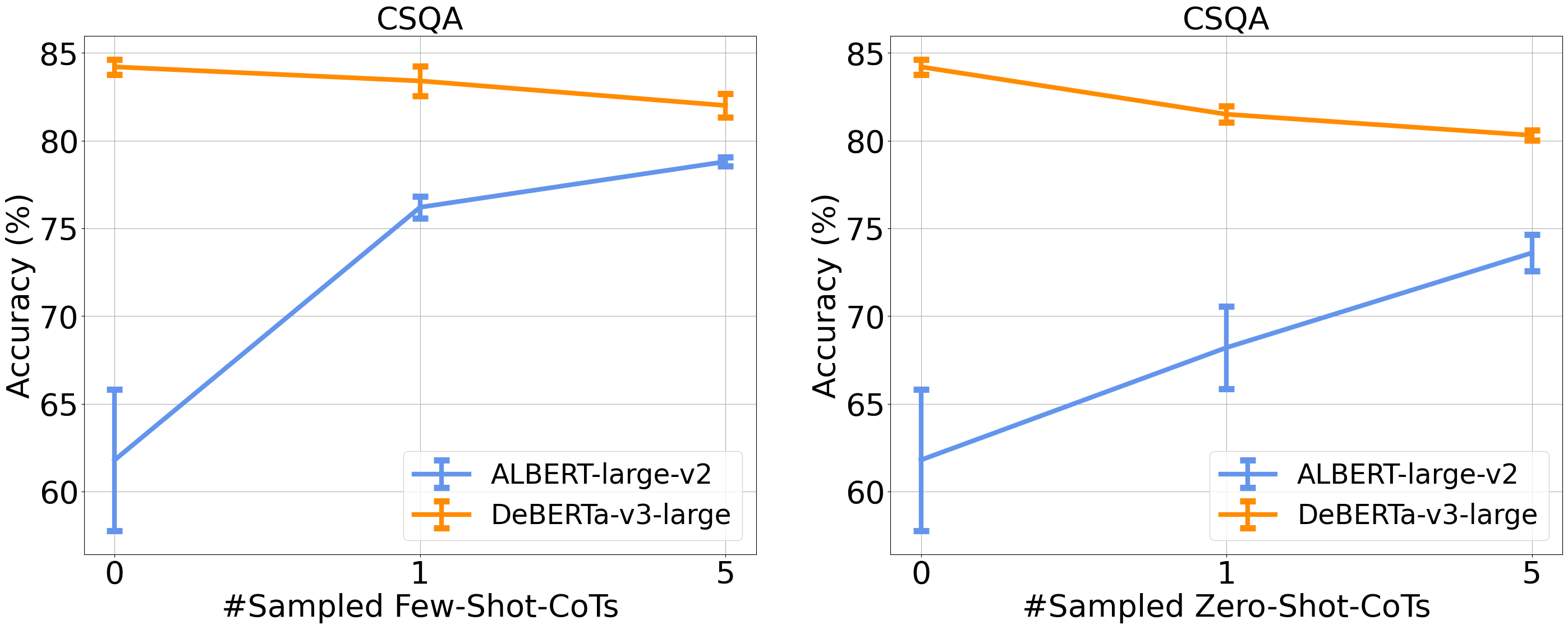}
    \caption{Accuracy of CSQA. Performance over various numbers of CoTs used in CoT-KA.}
    \label{figA:CSQA}
\end{figure*}

\begin{figure*}[htb]
    \centering
    \includegraphics[width=1 \linewidth]{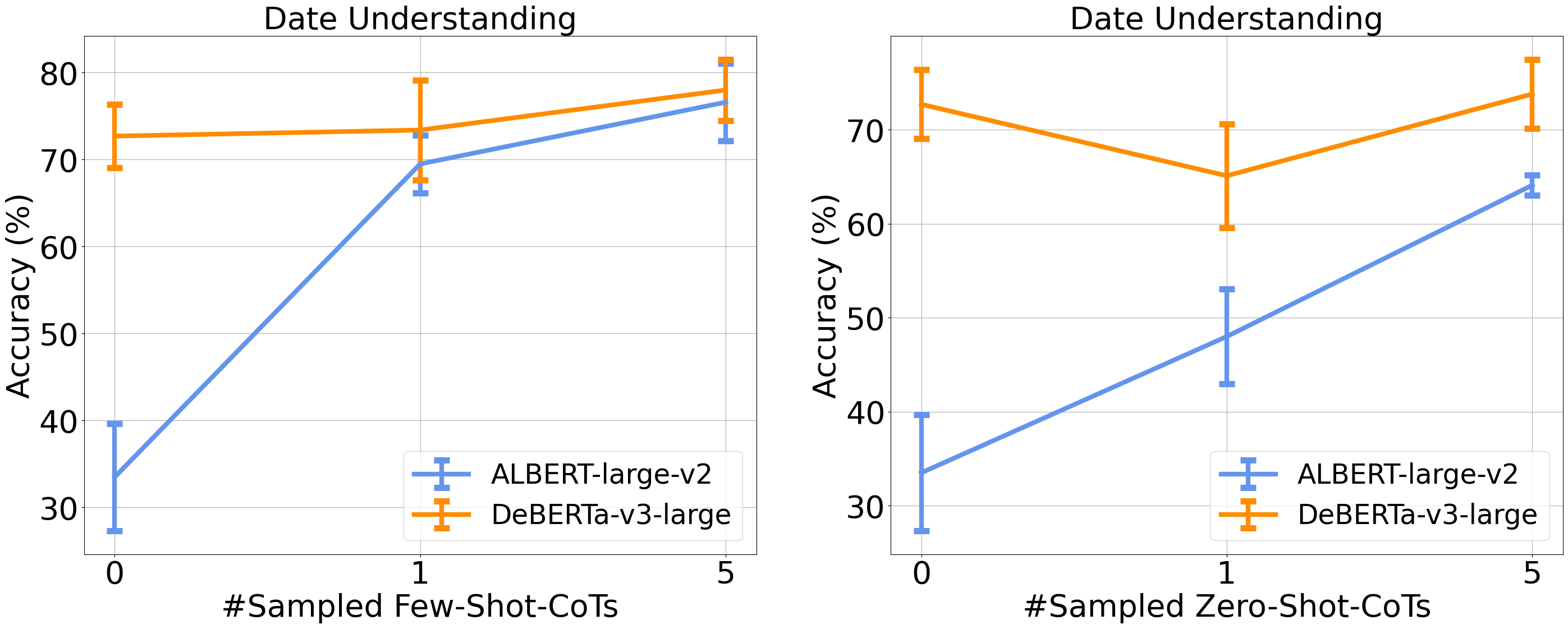}
    \caption{Accuracy of Date Understanding. Performance over various numbers of CoTs used in CoT-KA.}
    \label{figA:Date}
\end{figure*}

\begin{figure*}[htb]
    \centering
    \includegraphics[width=1 \linewidth]{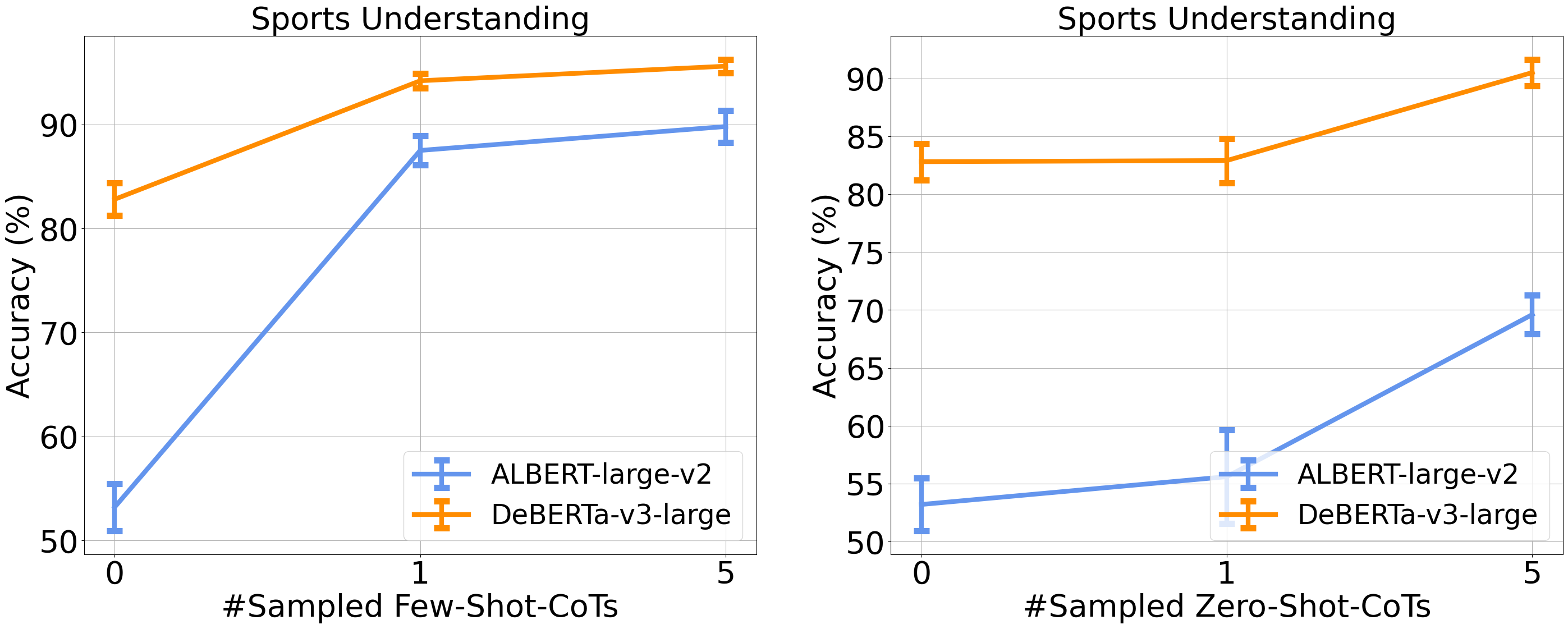}
    \caption{Accuracy of Sports Understanding. Performance over various numbers of CoTs used in CoT-KA.}
    \label{figA:Sports}
\end{figure*}

\begin{figure*}[htb]
    \centering
    \includegraphics[width=1 \linewidth]{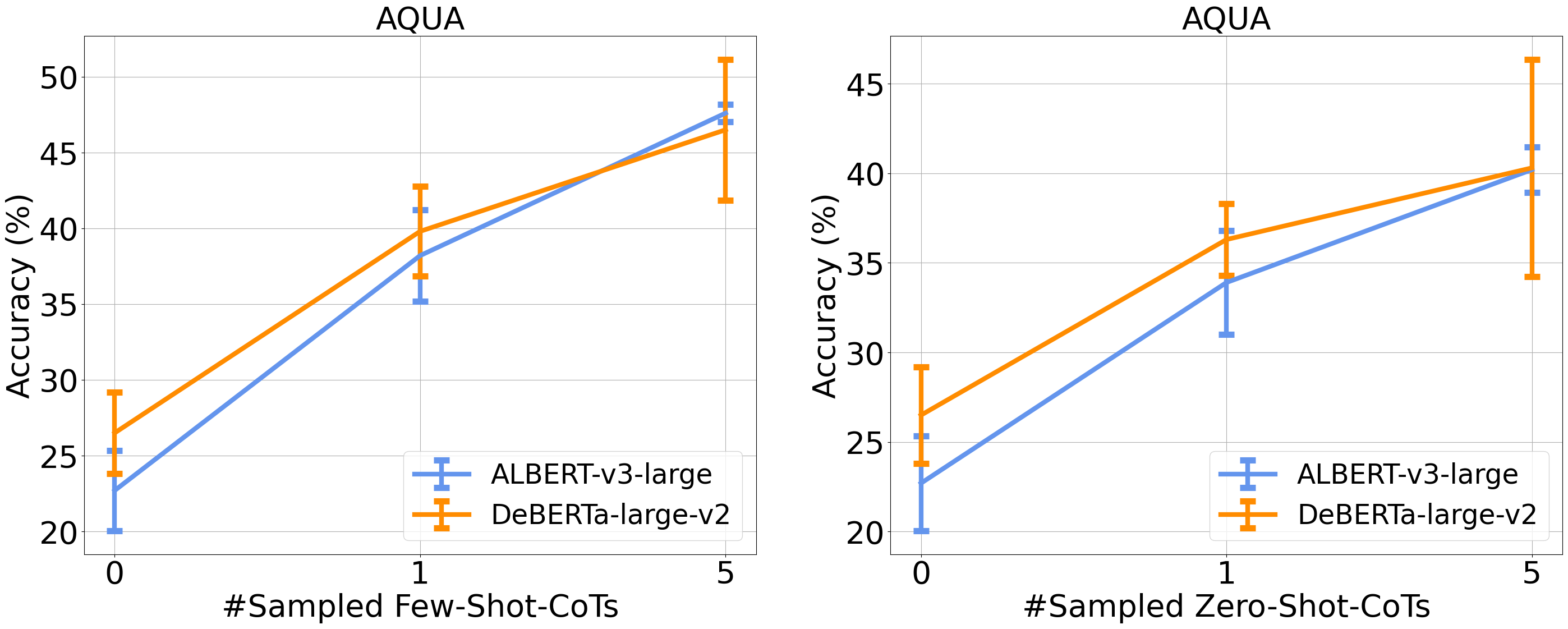}
    \caption{Accuracy of AQUA. Performance over various numbers of CoTs used in CoT-KA.}
    \label{figA:AQUA}
\end{figure*}

\begin{figure*}[htb]
    \centering
    \includegraphics[width=1 \linewidth]{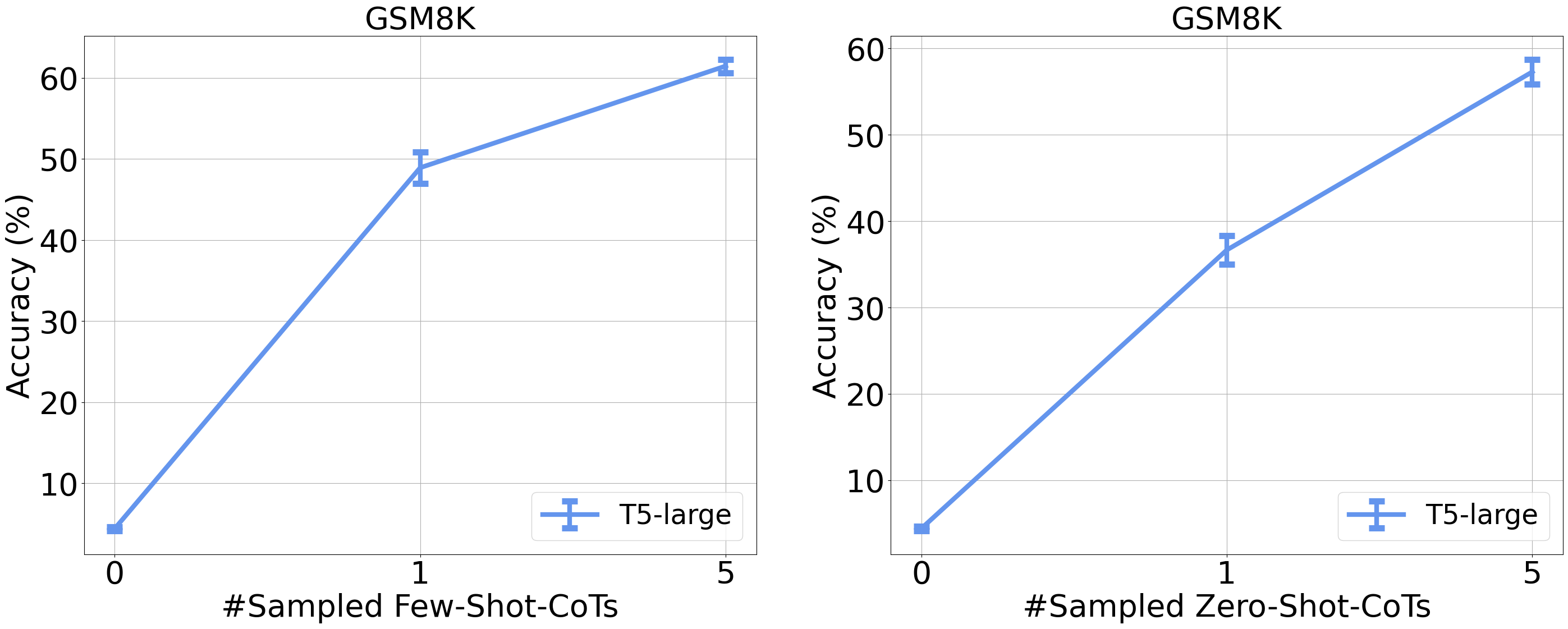}
    \caption{Accuracy of GSM8K. Performance over various numbers of CoTs used in CoT-KA.}
    \label{figA:GSM8K}
\end{figure*}

\begin{figure*}[htb]
    \centering
    \includegraphics[width=1 \linewidth]{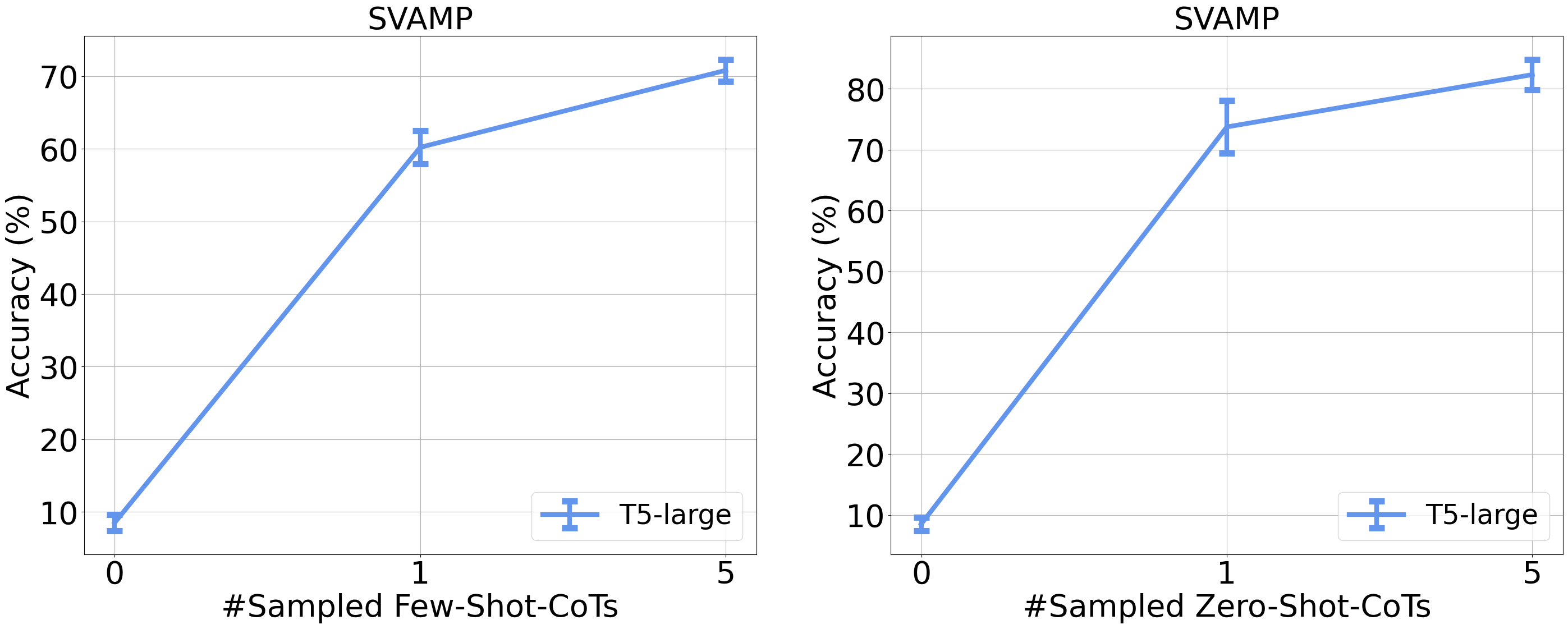}
    \caption{Accuracy of SVAMP. Performance over various numbers of CoTs used in CoT-KA.}
    \label{figA:SVAMP}
\end{figure*}

\begin{figure*}[htb]
    \centering
    \includegraphics[width=0.75 \linewidth]{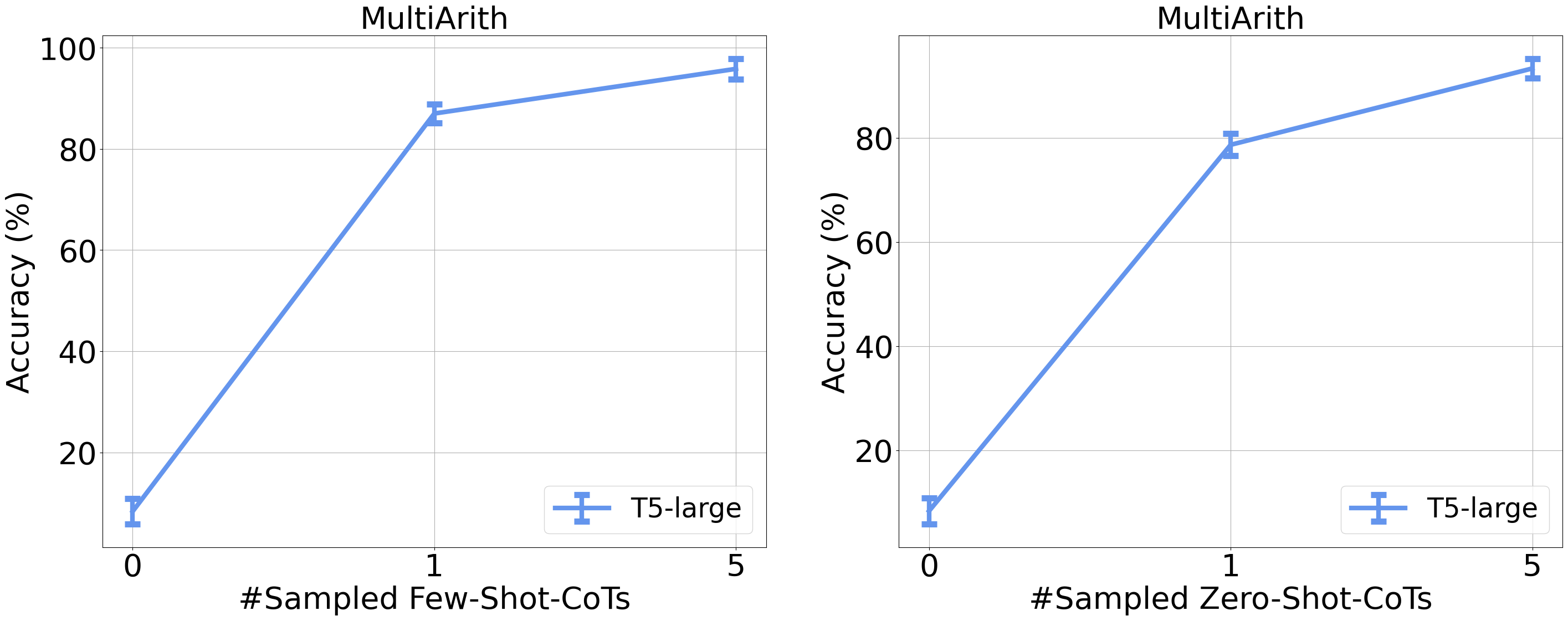}
    \caption{Accuracy of MultiArith. Performance over various numbers of CoTs used in CoT-KA.}
    \label{figA:MultiArith}
\end{figure*}

\begin{figure*}[htb]
    \centering
    \includegraphics[width=0.75 \linewidth]{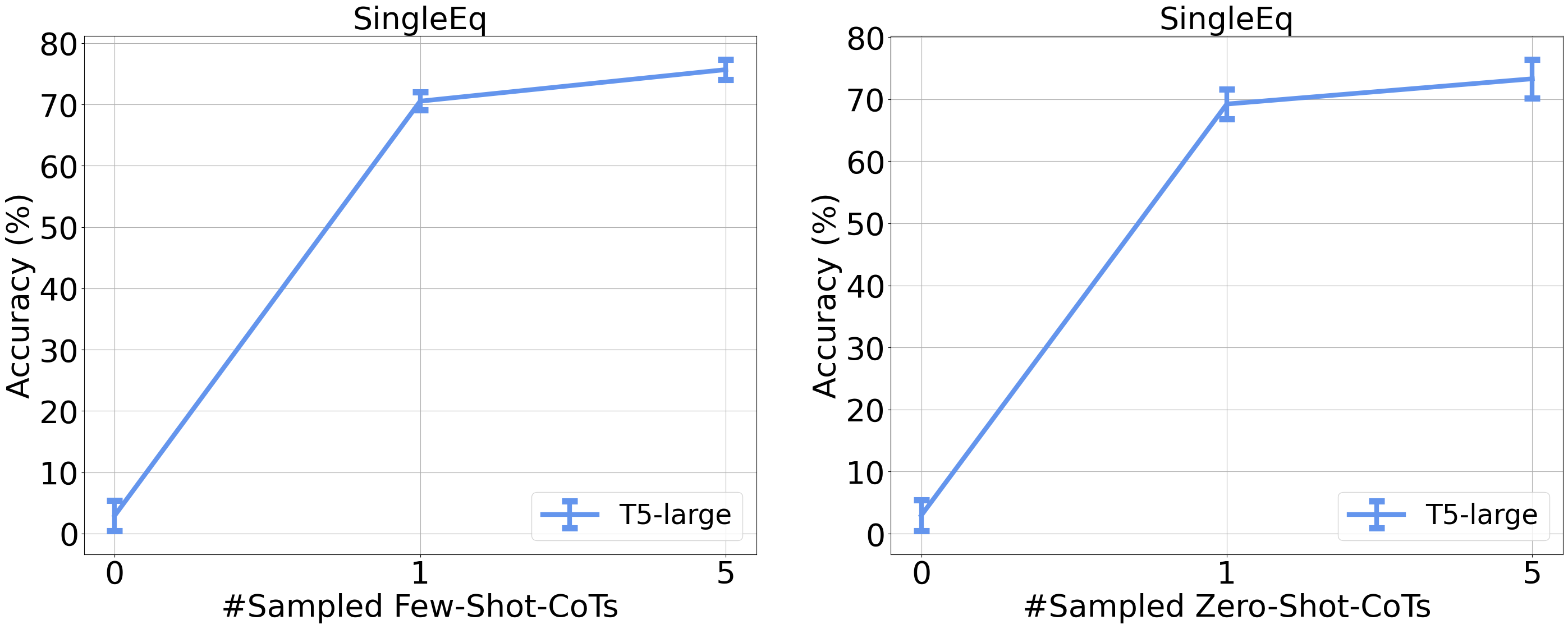}
    \caption{Accuracy of SingleEq. Performance over various numbers of CoTs used in CoT-KA.}
    \label{figA:SingleEq}
\end{figure*}

\begin{figure*}[htb]
    \centering
    \includegraphics[width=0.75 \linewidth]{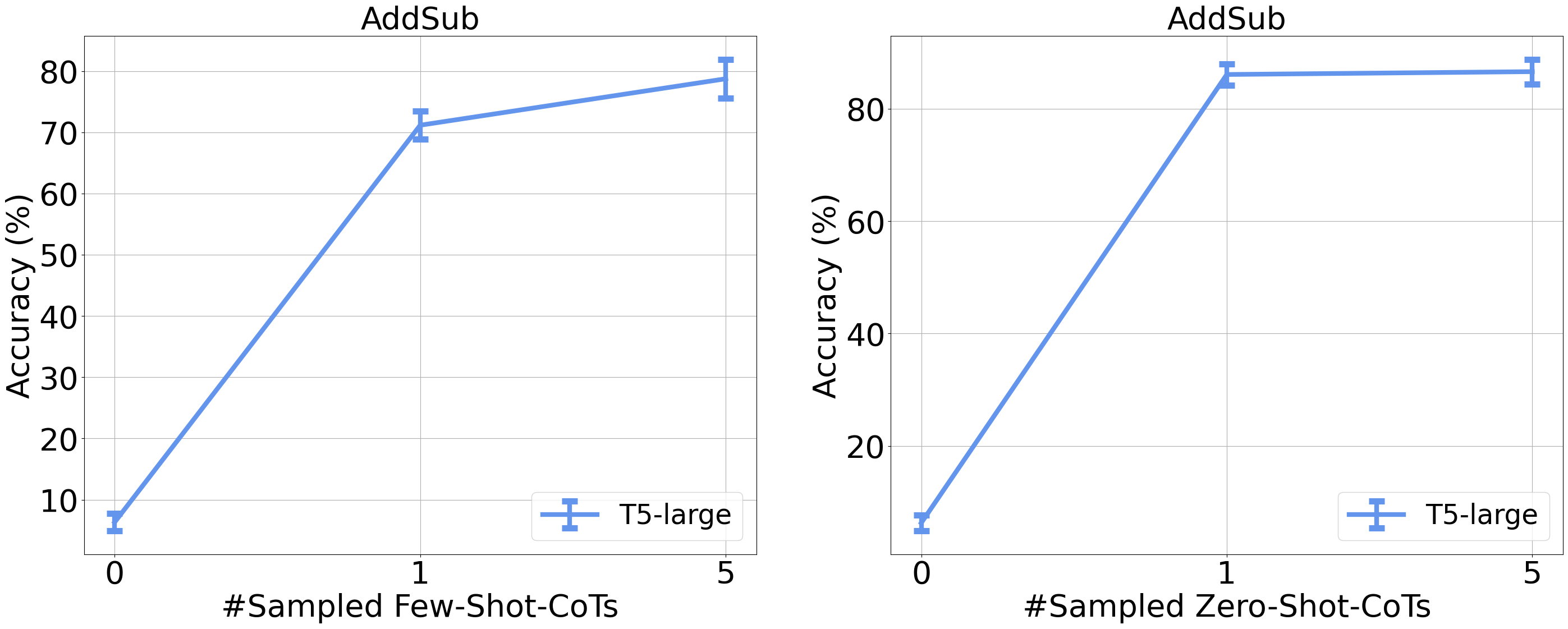}
    \caption{Accuracy of AddSub. Performance over various numbers of CoTs used in CoT-KA.}
    \label{figA:AddSub}
\end{figure*}

\begin{figure*}[htb]
    \centering
    \includegraphics[width=0.75 \linewidth]{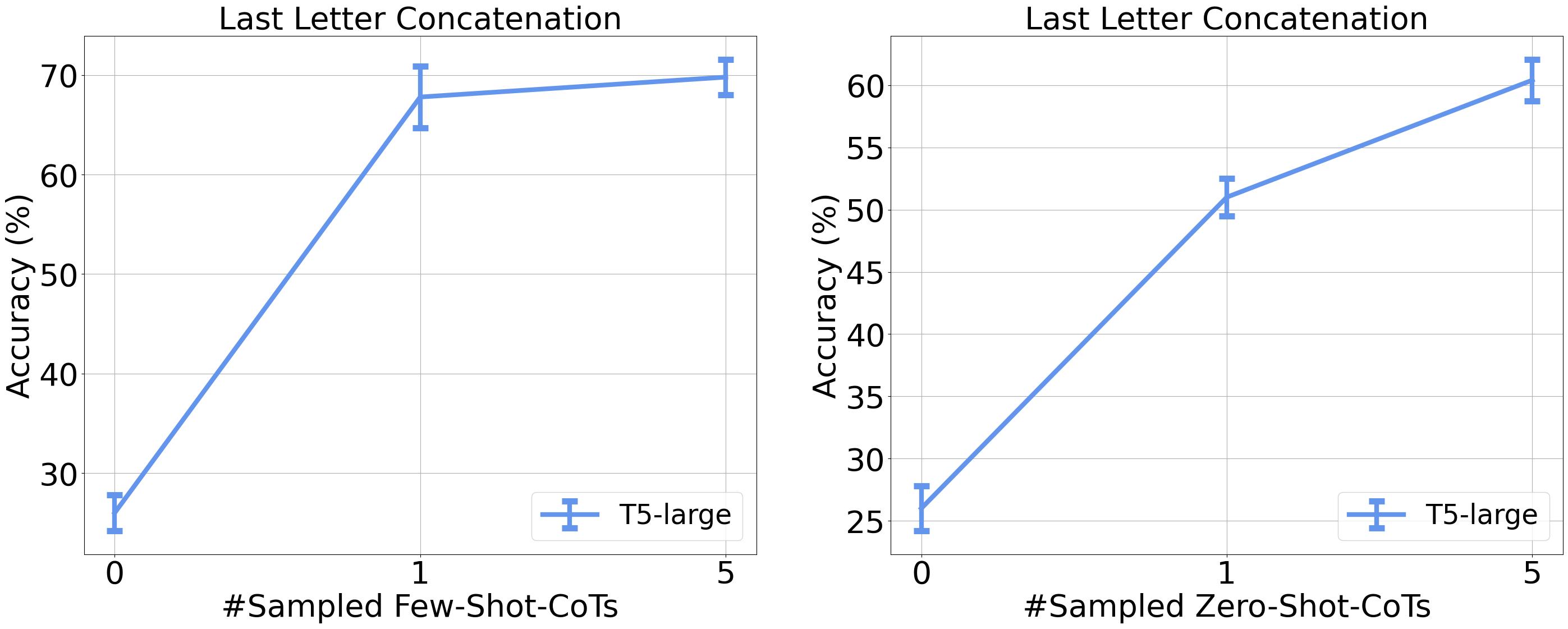}
    \caption{Accuracy of Last Letter Concatenation. Performance over various numbers of CoTs used in CoT-KA.}
    \label{figA:Letter}
\end{figure*}

We vary the number of sampled CoTs (1, 5) in \smodel and evaluate its performance on ten tasks, excluding StrategyQA. 
Figures from \ref{figA:CSQA} to \ref{figA:Letter} indicate that in most of these tasks, increasing the number of CoTs from 0 to 1 significantly improves task performance. 
However, when using DeBERTa-v3-large as the PLM, the performance gain in \smodel for CSQA, Date Understanding, and Sports Understanding is slight and even leads to a degradation. Furthermore, increasing the number of CoTs from 1 to 5 has a relatively small performance gain in \smodel (DeBERTa), except for improved Date Understanding and continued degradation in CSQA.

We observe that if the baseline, 
where the dataset is not augmented by a CoT, 
% without augmented by CoTs,
starts with a lower performance, the performance gain in \smodel becomes more significant as the number of CoTs increases.

\end{document}